\documentclass[review]{elsarticle}

\usepackage{lineno,hyperref,amsmath,multirow,epstopdf}
\usepackage{amssymb}		% assumes amsmath package installed
\usepackage{subfigure}		% multiple figures under the same caption

\modulolinenumbers[5]

\journal{Journal of \LaTeX\ Templates}

\begin{document}

\begin{frontmatter}

\title{Detection of bimanual gestures everywhere: why it matters, what we need and what is missing}

\author{Divya Shah\fnref{note:contrib}}
\author{Ernesto Denicia\fnref{note:contrib}}
\author{Tiago Pimentel\fnref{note:contrib}}
\fntext[note:contrib]{The authors contributed equally.}

\author{\\Barbara Bruno\fnref{mark:corresponding}}
\fntext[mark:corresponding]{Corresponding author. \textit{Email address}: \texttt{barbara.bruno@unige.it}}

\author{Fulvio Mastrogiovanni}

\address{Department of Informatics, Bioengineering, Robotics and Systems Engineering,\\ University of Genoa, via Opera Pia 13, 16145 Genoa, Italy.}

\begin{abstract}

Bimanual gestures are of the utmost importance for the study of motor coordination in humans and in everyday activities. 
A reliable detection of bimanual gestures in unconstrained environments is fundamental for their clinical study and to assess common activities of daily living.
This paper investigates techniques for a reliable, unconstrained detection and classification of bimanual gestures.
It assumes the availability of inertial data originating from the two hands/arms, builds upon a previously developed technique for gesture modelling based on Gaussian Mixture Modelling (GMM) and Gaussian Mixture Regression (GMR), and compares different modelling and classification techniques, which are based on a number of assumptions inspired by literature about how bimanual gestures are represented and modelled in the brain. 
Experiments show results related to $5$ everyday bimanual activities, which have been selected on the basis of three main parameters: (not) constraining the two hands by a physical tool, (not) requiring a specific sequence of single-hand gestures, being recursive (or not). 
In the best performing combination of modelling approach and classification technique, five out of five activities are recognised up to an accuracy of 97\%, a precision of 82\% and a level of recall of 100\%.
\end{abstract}


\end{frontmatter}

%\linenumbers

\section{Introduction}

Bimanual gestures are central to everyday life, and constitute a fundamental ground for the study of basic principles of human behaviour.
Traditionally, the study of bimanual gestures in humans focus on very simple motions involving fingers and hands, and including coordination, symmetry, in-phase and anti-phase behaviours.
These studies are aimed at understanding the dynamics associated with bimanual movements and target aspects of motor control, as exemplified by the Haken-Kelso-Bunz (HKB) model for self-organisation in human movement pattern formation \cite{Kelso1984}\cite{Hakenetal1985}.

In this paper, we are interested in determining how the study of bimanual gestures can lead to automated systems for their detection and classification in unconstrained, everyday environments.
In the context of assistive systems for fragile populations, including elderly, people with disabilities and other people with special needs, the need arises to provide caregivers, medical staff or simply relatives with a tool able to assess the ability of assisted people to perform bimanual gestures in their \textit{natural} environment. 
Such approach is in line with the \textit{ageing in place} paradigm, a recent healthcare position which acknowledges and focuses on the role that a person's surroundings (the home, the neighbourhood) play for his well-being in older ages \cite{Wiles11}. 
A familiar environment brings about a sense of security, independence and autonomy which has a positive impact on routines and activities, and ultimately on quality of life.

There is a big gap between clinical studies involving the coordination of finger movements and the recognition of such general-purpose bimanual gestures as \textit{opening and closing curtains}, \textit{sweeping the floor}, or \textit{filling a cup with water}.  
However, the first step to take is to determine how current understanding of bimanual movements and their representation in the brain can lead to better detection and classification systems in real-world environments.
Three factors must be considered when designing such a system.

\textit{Factor 1. It is debated whether bimanual gestures are controlled in intrinsic or extrinsic coordinates, or rather multiple coordination strategies co-occur}. 

Bimanual movements tend to motion symmetry and stabilisation \cite{Kelso1984}.
This has been typically explained using co-activation of homologous muscles in neuronal motor structures, due to bilateral cross-talk, suggesting that bimanual coordination is mostly done using intrinsic (i.e., proprioceptive) coordinates.
Mechsner \textit{et al} suggest that, instead, such a coordination is due to spatial, perceptual symmetry only, i.e., using extrinsic (i.e., exteroceptive) visual coordinates \cite{Mechsneretal2001}. 
If this were true, there would be no need to map visual representations to motor representations (and \textit{viceversa}), and voluntary movements could be organised on the basis of perceptual goals.
The role of different coordinates and their interplay for bimanual coordination mechanisms has been studied by Sakurada and colleagues \cite{Sakuradaetal2015}.
Starting from studies relating temporal and spatial couplings in bilateral motions, including the adaptation exhibited by two hands having to perform motions of different speed (i.e., the fastest becoming slightly slower and viceversa) \cite{Heueretal1998}, and the fact that the movement of a non-dominant hand is likely to be assimilated by the dominant one \cite{Byblowetal2000}, they demonstrate a relatively stronger contribution of intrinsic components in bimanual coordination, although both components are flexibly regulated according to the specific bimanual task.
Furthermore, they argue that the central nervous system regulates bilateral coordination at different levels, as hypothesised by Swinnen and Wenderoth \cite{SwinnenandWenderoth2004}.
The importance of both intrinsic and extrinsic coordinates seems to be confirmed by recent studies in interpersonal coordination \cite{Kodamaetal2015}.
It is suggested that coordinated motion is informed by a full perception-action coupling, including visual and haptic sensorimotor loops, which propagates to the neuromuscular system.

We derive two requirements for our analysis:
\begin{itemize}
\item[$R_1$)] we must consider models \textit{agnostic} with respect to an explicit coordination at the motor level between the two hands/arms;
\item[$R_2$)] classification techniques must be robust to variation in speed, both for the bimanual gesture as a whole and for the single hand/arm. 
\end{itemize} 

\textit{Factor 2. Ageing affects the way we move, and therefore coordination in bimanual gestures varies over time}.

According to the \textit{dedifferentiation} paradigm, ageing is now considered a parallel and distributed process occurring at various levels in the human's body.
The dedifferentiation can be defined as ``the process by which structures, mechanisms of behaviour that were specialised for a given function lose their specialisation and become simplified, less distinct or common to different functions'' \cite{BaltesandLindenberger1997}.
As a consequence, ageing affects not only individual body \textit{subsystems} (i.e., the muscular system or the brain), but also their interactions.
It is argued by Sleimen-Malkoun and colleagues that such a process can lead to common and intertwined causes for \textit{cognitive ageing}, i.e., a general slowing down of information processing, including the information related to procedural memory and -- therefore -- movement and coordination \cite{SleimenMalkounetal2014}.
It is posited that the ageing brain undergoes anatomical and physiological changes, for the reorganising activation patterns between neural circuits.
As far as motor task complexity is concerned, a generalised increased activation of brain areas is even more evident, which reflects a greater involvement of processes related to executive control.

Also in this case, we derive an important requirement:
\begin{itemize}
\item[$R_3$)] we must consider models which can be adapted over time and which follow the evolution of the musculoskeletal system, at least implicitly, thus requiring the use of forms of machine learning techniques.
\end{itemize}  

\textit{Factor 3. Different mental representations of sensorimotor loops and action, involving discrete and continuous organisation principles, are under debate}.

Beside the models aimed at representing bimanual gestures assuming a motor control framework, much work has been carried out in the past few years to devise building blocks for mental action representation \cite{Kelso1984}\cite{Hakenetal1985}\cite{Mechsneretal2001}\cite{SwinnenandWenderoth2004}\cite{Sakuradaetal2015}\cite{Kodamaetal2015}.
Assuming a goal-directed cognitive perspective, it has been shown how movements can be represented as a serial and functional combination of goal-related body postures, or goal postures (i.e., key frames), as well as their transitional states.
Furthermore, it has been posited that movements can be expressed as incremental changes between goal postures, which reduces the amount of effortful attention needed for their execution \cite{Rosenbaumetal2007}.
On these premises, Basic Action Concepts (BACs) have been proposed as building blocks of mental action representations.
BACs represent chucked body postures related to common functions to realise goal-directed actions.
Schack and colleagues posit that complex (including bimanual) actions are mentally represented as a combination of executed action and intended or observed effects \cite{Schacketal2014}.
Furthermore, they argue that the map linking motion and perceptual effects is bi-directional and stored hierarchically in long-term memory, in structures resembling \textit{dendograms} \cite{Schack2012}.
This is a specific case of what has been defined by Bernstein as the \textit{degrees of freedom problem} \cite{Bernstein1967}.
The problem is related to how the various parts of the motor system can become harnessed so as to generate coordinated behaviour when needed.
As Bernstein theorised, a key role is played by muscular-articular links (i.e., synergies) in constraining how many degrees of freedom lead to dexterous behaviour.
Harrison and Stergiou argue that dexterity and motion robustness are enabled by multi-functional (degenerate) body parts able to assume context-dependent roles.
As a consequence, task-specific human-environment interactions can flexibly generate adaptable motor solutions.

We derive two requirements:
\begin{itemize}
\item[$R_4$)] although motion models are intrinsically continuous, we need to derive a discrete representation able to provide action labels which, in principle, can lead to more complex organisations;
\item[$R_5$)] models must capture dexterity in everyday environments and robustness to different executions of the same gesture, which leads to models obtained by human demonstration.
\end{itemize}

On the basis of these requirements, we propose a bimanual wearable system able to detect and classify bimanual gestures using the inertial information provided by two wrist-mounted smartwatches.
The system builds up and significantly extends previous work \cite{Bruno12}, and adheres to the \textit{wearable sensing} paradigm, which envisions the use of sensors located on a person's body, either with wearable devices such as smartwatches, or with purposely engineered articles of clothing \cite{Lara13}, to determine a number of important parameters, in our case motion.
Since sensors are physically carried around, the monitoring activity can virtually occur in any place and it is usually focused on the detection of movements and gestures.

The contribution is four-fold: (i) an analysis of two procedures for modelling bimanual gestures, respectively explicitly and implicitly taking the correlation between the two hands/arms into account; (ii) an analysis of two procedures for the classification of run-time data, respectively relying on the probability measure and the Mahalanobis distance to compute the similarity between run-time data and previously stored models of bimanual gestures; (iii) a performance assessment of the developed techniques with the standard statistical metrics of accuracy, precision and recall over the collected dataset, as well as under real-life conditions; (iv) a dataset of $60$ recordings of five bimanual gestures, performed by ten volunteers, to support reproducible research.

The paper is structured as follows.
Section \ref{sec:related_work} describes the theoretical background of the proposed modelling and recognition procedures, as well as related work, in view of the requirements outlined above.
Section \ref{sec:system_architecture} provides a thorough description of the system's architecture and insights on the five bimanual gestures considered for the analysis; the performance of such system are presented and discussed in Section \ref{sec:experimental_evaluation}.
Conclusions follow.

\section{Related Work}
\label{sec:related_work}

Wearable systems for the automatic recognition of human gestures and full-body movements typically rely on inertial information.
Accelerometers prove to be the most informative sensor for the task \cite{Lester05}.
To comply with end users' constraints related to the impact of the monitoring system on their appearance and freedom of motion, most solutions adopt a single sensing device, either located at the waist \cite{Mathie04} or, as it is becoming more and more common, at the wrist \cite{Dietrich14}.

\begin{figure}
\centering
\includegraphics[width=11cm]{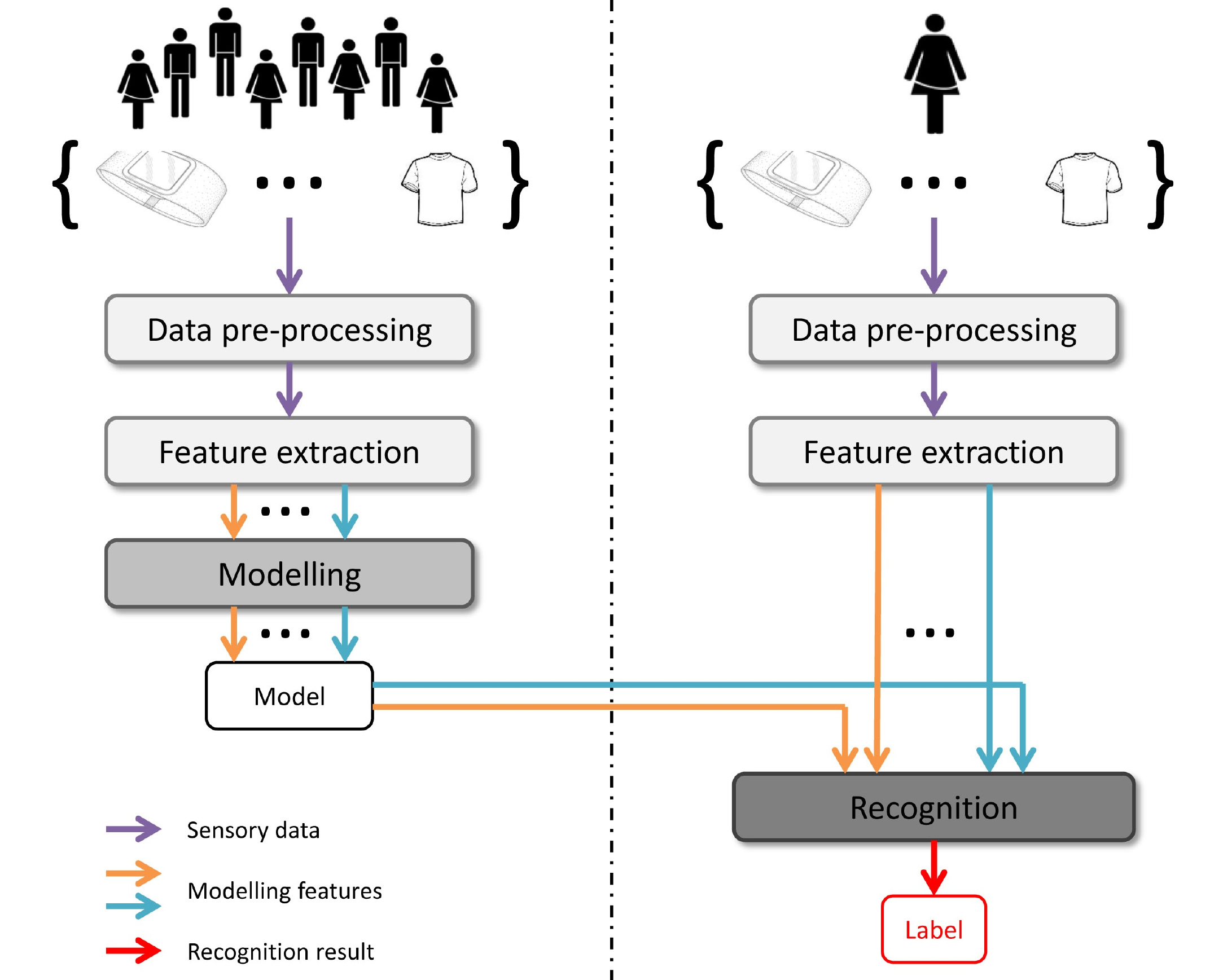}
\caption{The typical architecture of wearable sensing systems for the recognition of human gestures. The left hand side lists the tasks of the training phase, while the right hand side lists the tasks executed during the testing phase.}
\label{fig:system_architecture}
\end{figure}

Due to the similarities in the input data and in the operating conditions, most systems adopt a similar architecture \cite{Lara13}, sketched in Figure \ref{fig:system_architecture}.
The architecture identifies two stages, namely a training phase (on the left hand side in the Figure) and a testing phase (on the right hand side).
The \textit{training} phase, typically executed off-line without strict computational constraints, is devoted to the creation of a compact representation of a gesture/movement of interest, on the basis of an informative set of examples.
This also complies with requirements $R_3$ and $R_5$ above.
The \textit{testing} phase, which may be subject to real-time and computational constraints, is responsible for the analysis of an input data stream to detect the gesture, among the modelled ones, which more closely matches it, if any, and label it accordingly.
Please note that the word ``testing'' is used here with respect to the data stream to analyse, with no reference to the stage of development of the monitoring system.
Specifically, we denote with the term ``validation'' the development stage in which we assess the performance of the system, and with the term ``deployment'' the stage in which the system is actually adopted by end users.
During validation, the testing phase executes an off-line analysis of labelled gesture recordings, while during deployment the testing phase executes an on-line analysis of a continuous data stream, in unsupervised conditions.

During the training phase (see Figure \ref{fig:system_architecture} on the left hand side), it is first necessary to acquire and build the training set of measured attributes for the gestures of interest.
Two approaches are possible.
The \textit{specialised} approach envisions the creation of a training set exclusively composed of gesture recordings performed by the person to be monitored during the deployment stage.
This approach maximises the recognition accuracy at the expenses of a long setup for each new installation.
However, it enforces requirements $R_3$ and $R_5$, in that it allows someone to periodically retrain the system if necessary.
Conversely, the \textit{generalised} approach envisions the creation of a training set composed of a large number of gesture recordings, performed by a number of volunteers (not necessarily including the person to monitor).
Using gestures provided by different individuals maximises the likelihood that the model is able to capture a much varied dexterity, as posited by $R_5$. 
This approach, albeit more prone to errors, greatly reduces the setup costs and is to be preferred in the case of Ambient Assisted Living applications, in which the perceived system's ease of use is crucial for its success \cite{Bruno12, Bulling14}.

Once the training set is available, it is typically filtered for noise reduction and/or formatted for later processing (\textit{data pre-processing} stage).
Then, the purpose of the \textit{feature extraction} procedure is to determine relevant information (in the form of features) from raw signals.
Features are expected to (i) enhance the differences between gestures while being invariant when extracted from data patterns corresponding to the same gesture, (ii) lead to a compact representation of the gesture and (iii) require limited computational time and resources for the extraction and processing, since these operations are subject to real-time constraints during deployment.
Literature discriminates between \textit{statistical} features, extracted using methods such as the Fourier and the Wavelet transform on the basis of quantitative characteristics of the data, and \textit{structural} features, aiming at enhancing the interrelationship among data \cite{Lara13}.
Most human activity recognition systems based on inertial measurements make use of statistical features, usually defined in the time- or frequency-domain \cite{Krassnig10}.
Alternative feature extraction methods include the use of Principal Component Analysis \cite{Mashita12} and autoregressive models \cite{Lee11}.
With respect to time-domain features exclusively, which minimise the computational load introduced by the feature extraction procedure, gravity and body acceleration are among the most commonly adopted \cite{Karantonis06, Krassnig10, Bruno12}.
Discriminating between gravity and body acceleration is a non-trivial operation, for the two features overlap both in the time and frequency domains.
Two approaches are typically adopted to separate them.
The former exploits additional sensors, such as gyroscopes \cite{Chul09} or magnetometers \cite{Bonnet07}, to compute the orientation or attitude of another body part, usually the torso.
The latter exploits the known features of gravity and uses either low-pass filters to isolate the gravitational component \cite{Karantonis06, Bruno12}, or high-pass filters to isolate the body acceleration components \cite{Sharma08}.

Once gravity and body acceleration components are isolated, the need arises to model them as features.
Six different 1-dimensional sampled signals are available, i.e., the three $g_x, g_y, g_z$ gravity and the three $b_x, b_y, b_z$ body acceleration components along the $x$, $y$ and $z$ axes.
Again, two possibilities are discussed in the Literature.
The first is to assume the signals to be pairwise uncorrelated, which yields six separate 2-dimensional features, each feature being composed of timing information and the corresponding signal value on a given axis, i.e., $(t, g_i)$ and $(t, b_i)$, where $i \in \{x, y, z\}$.
The second is to assume the $x$, $y$ and $z$ components of gravity and body acceleration to be correlated, which yields two separate 4-dimensional features, i.e., $(t, g_x, g_y, g_z)$, $(t, b_x, b_y, b_z)$, each feature being composed of timing information and the corresponding signal values on all axes.
The explicit use of correlation among tri-axial acceleration data has been proved to lead to better results in terms of classification rate and computational time \cite{Cho08, Krassnig10, Bruno12}.
It is noteworthy that, in case of bimanual gestures, it is possible to explicitly model the correlation among inertial data originating from the two hands/arms, or to consider them as separate signals.
In this way, we can comply with requirements $R_1$ and (in part) $R_2$ above. 

Finally, the \textit{modelling} procedure is devoted to the creation of a compact and informative representation of the considered gestures in terms of available sensory data.
Two classes of approaches have been traditionally pursued.
In \textit{logic-based} solutions each gesture to monitor and recognise is encoded through sound and well-defined rules, which are based on ranges of admissible values for a set of relevant parameters.
Recognition is carried out by analysing run-time sensory values to progressively converge towards the encoded gesture more closely matching run-time data.
Decision trees, which allow for a fast and simple classification procedure, are the most adopted solution in logic-based approaches \cite{Lee02, Mathie04, Karantonis06, Krassnig10}.
\textit{Probability-based} solutions assume instead each gesture to be represented by a model encoding relevant moments of the training set, and to be identified using non ambiguous labels.
In this case, recognition is typically performed by comparing run-time sensory data with the stored models through probabilistic distance measures.
Commonly adopted techniques include Neural Networks \cite{Krassnig10}, Hidden Markov Models \cite{Minnen05, OlguinOlguin06, Choudhury08} and Gaussian Mixture Models \cite{Bruno12}.
In our work, we exploit probability-based models to comply with requirement $R_4$.

During the testing phase (see Figure \ref{fig:system_architecture} on the right hand side), analogously to what happens in the training phase, a number of steps are sequentially executed.
The feature extraction step executes the same algorithms of the training phase. Once the testing stream has been processed (typically focusing on a time window), it is possible to evaluate its features against the previously trained models (\textit{recognition}), generating a predicted label.
In the testing phase, we exploit specific distance metrics relating the stored models with the run-time data stream. 
In this way, we can account for requirement $R_2$.

Most wearable sensing systems based on a single sensing point (e.g., the right wrist) focus on the recognition of gestures which are either one-handed, e.g., \textit{bringing a cigarette to the lips to smoke} \cite{Dietrich14}, or, albeit involving both hands, such as \textit{cutting meat with fork and knife} \cite{Bruno13} or even the full-body, e.g., \textit{climbing stairs} \cite{Bruno12}, correspond to a unique and generalised pattern at the considered sensing point.
The presented work relaxes this assumption, by evaluating a wearable sensing system based on two sensing points (the left and right wrists) which allows for the modelling and recognition of \textit{generic} bimanual gestures.

\section{System Architecture}
\label{sec:system_architecture}

\begin{figure}
\centering
\includegraphics[width=10.8cm]{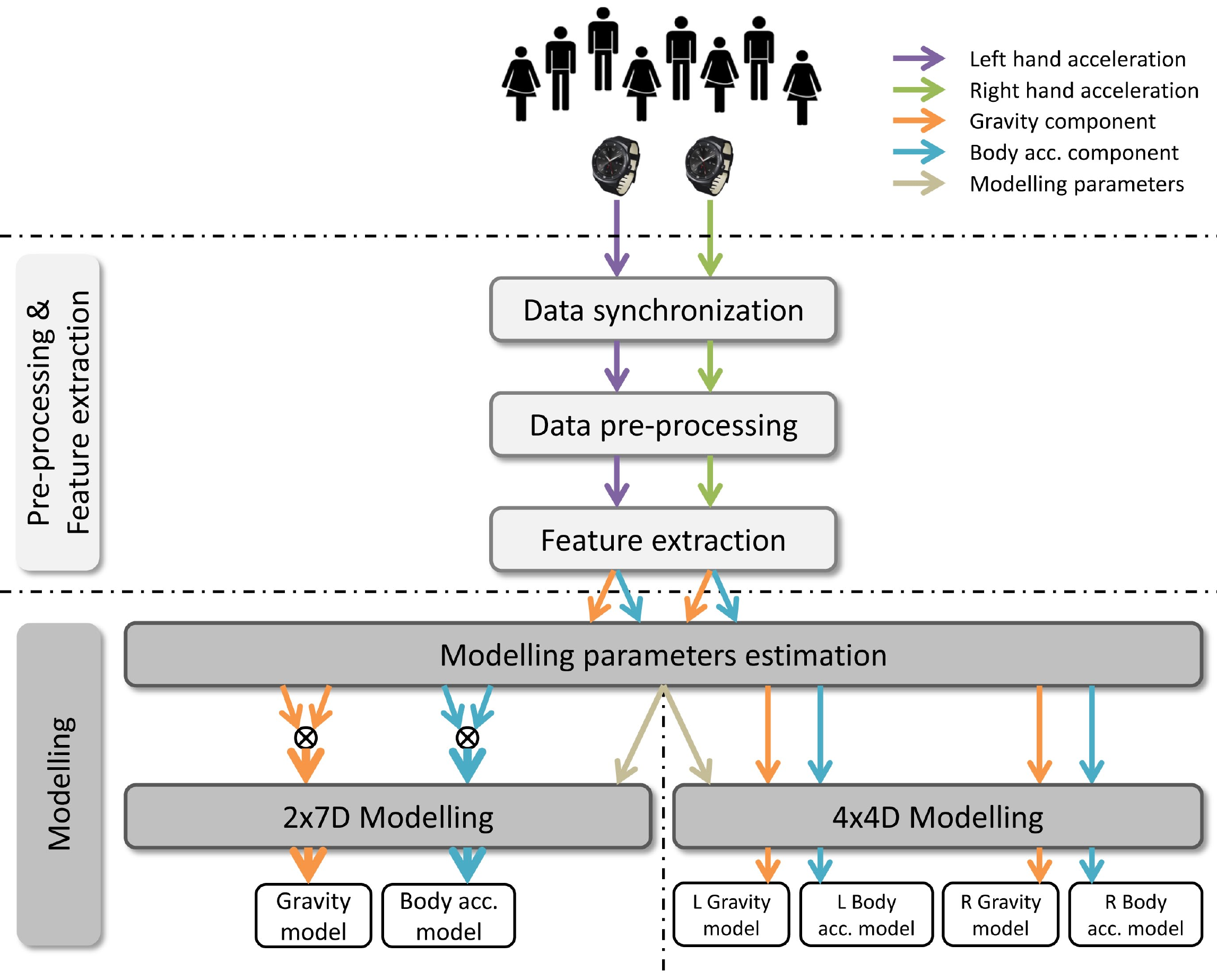}
\caption{System architecture (training phase).}
\label{fig:modelling_architecture}
\end{figure}
Figure \ref{fig:modelling_architecture} shows a schematic view of the training phase of the proposed system, while Figure \ref{fig:testing_architecture} details the operations performed during the testing phase.
The blocks devoted to \textit{data synchronisation} and \textit{data pre-processing}, as well as the \textit{feature extraction} block, are the same in the two phases.
We consider two approaches for the modelling stage: (i) \textit{explicit} modelling of the correlation of the two hands ($2\times7D$ approach, see Figure \ref{fig:modelling_architecture} on the left hand side), presupposing the stress on intrinsic coordinates in motor control studies, and (ii) \textit{implicit} modelling of the correlation ($4\times4D$ approach, see Figure \ref{fig:modelling_architecture} on the right hand side), which assumes the importance of extrinsic coordinates.
We also consider two approaches for the comparison of testing data with the available models, respectively based on the probability measure and the Mahalanobis distance.
The former takes into account the likelihood of a model \textit{as a whole}, whereas the second weights more the contribution of body degeneracy in robustness and dexterity.

\subsection{Pre-processing \& Feature extraction}

The proposed system relies on the inertial information collected by two tri-axial accelerometers, respectively located at a person's left and right wrists.
To properly manage the two data streams, they should be synchronised, and the \textit{data synchronisation} procedure heavily depends on the choices made related to hardware solutions.
As it will be described in Section \ref{sec:experimental_evaluation}, in the case of the devices we adopt, the synchronisation is guaranteed by the manufacturer.

All the trials of a gesture in the training set (i.e., all the couples of left- and right-wrist data streams associated with a single execution of the gesture) are initially synchronised with each other manually, so that the starting moment of the gesture is the same across all recordings, and trimmed to have equal length.
Then, the \textit{data pre-processing} stage filters each acceleration stream with a median filter of size $3$ to reduce noise.

Let us assume that we have $M$ different bimanual gestures.
For each gesture $m$ to learn, where $m=1, \ldots, M$, let us assume that the training set includes $S_m$ trials and $s$ is one of them. After the pre-processing stage, all the $S_m$ trials are synchronised and truncated as to be composed of the same number $K_m$ of observations.
A trial is defined as:
\begin{equation}
s = \{g_{l,k}, b_{l,k}, g_{r,k}, b_{r,k}\} \quad k = 1, \ldots, K_m
\label{eq:4Dtrial}
\end{equation}
with:
\begin{equation}
\begin{split}
g_{l,k}&=(t_k, g_{l,x,k}, g_{l,y,k}, g_{l,z,k}), \\
b_{l,k}&=(t_k, b_{l,x,k}, b_{l,y,k}, b_{l,z,k}), \\
g_{r,k}&=(t_k, g_{r,x,k}, g_{r,y,k}, g_{r,z,k}), \\
b_{r,k}&=(t_k, b_{r,x,k}, b_{r,y,k}, b_{r,z,k}), \\
\end{split}
\label{eq:4Dfeatures}
\end{equation}
where $l$ and $r$ denote, respectively, the acceleration streams provided by the sensing device on the left and on the right wrist, $g_{l,k}$ includes the time and $x, y$ and $z$ components of the gravity on the left acceleration stream, $b_{l,k}$ includes the time and $x, y$ and $z$ components of the body acceleration on the left acceleration stream, $g_{r,k}$ includes the time and $x, y$ and $z$ components of the gravity on the right acceleration stream and $b_{r,k}$ includes the time and $x, y$ and $z$ components of the body acceleration on the right acceleration stream.
The \textit{feature extraction} stage separates the $l$ and $r$ tri-axial acceleration streams provided by the sensing devices into their gravity and body acceleration components \cite{Bruno12}, by applying a low pass Chebyshev I $5^{\circ}$ order filter ($F_{cut}=0.25 {Hz}, A_{pass}= 0.001 {dB}, A_{stop}= -100 {dB},F_{stop}= 2 {Hz}$).

\subsection{Modelling}

The goal of the \textit{modelling} stage is to combine the $S_m$ trials in the training set to obtain a generalised version, i.e., a \textit{model}, of gesture $m$, defined in terms of the two features of gravity and body acceleration.
Two approaches are possible.

In the \textit{explicit} correlation modelling ($2\times7D$ approach, see Figure \ref{fig:modelling_architecture} on the left hand side), we merge the left and right components of each trial $s$ to create $7$-dimensional features, i.e.:
\begin{equation}
s = \{G_{k}, B_{k}\} \quad k = 1, \ldots, K_m
\label{eq:7Dtrial}
\end{equation}
with:
\begin{equation}
\begin{split}
G_{k} &= (t_k, g_{l,x,k}, g_{l,y,k}, g_{l,z,k}, g_{r,x,k}, g_{r,y,k}, g_{r,z,k}), \\
B_{k} &= (t_k, b_{l,x,k}, b_{l,y,k}, b_{l,z,k}, b_{r,x,k}, b_{r,y,k}, b_{r,z,k}), \\
\end{split}
\label{eq:7Dfeatures}
\end{equation}
and the model of gesture $m$ is then defined in terms of $G$ and $B$.

Conversely, in the \textit{implicit} correlation modelling ($4\times4D$ approach, see Figure \ref{fig:modelling_architecture} on the right hand side), we keep the left and right hand streams independent, thus considering the four features defined in \eqref{eq:4Dfeatures}.
The model of gesture $m$ is then defined in terms of $g_l, b_l, g_r$ and $b_r$.

The first approach corresponds to assuming that the motion of the two hands is \textit{fully constrained} by the performed gesture, while the latter approach leaves to later stages the responsibility of correlating the two data streams.
At the same time, in the first case we assume the contribution of the left and right hands/arms as correlated, whereas in the second case we do not pose such an assumption.
Albeit introducing an additional step, the possibility of tuning the correlation of the two data streams opens interesting scenarios for the recognition stage.
Consider the gesture of rotating a tap's handle with the left hand, which can occur in a number of situations (for example, when filling a glass with tap water, or when washing a dish): by varying the importance given to this hand we can either have a more flexible system, which is able to recognise many situations in light of the common traits in the left hand stream, or a more specialised one, which is focused on one situation only and is able to filter out all the others in light of the differences in the right hand stream.

The modelling procedure in itself is the same for both approaches and, in particular, it relies on Gaussian Mixture Modelling (GMM) and Gaussian Mixture Regression (GMR) for the retrieval of the \textit{expected} curve and covariance matrix of each considered feature, on the basis of a given training set. The procedure has been first introduced in the field of Human-Robot Interaction \citep{Calinon10} and later used for the purposes of Human Activity Recognition with a wrist-placed inertial device for a single arm \cite{Bruno12}. We point to the references for its thorough description.

We denote with $f$ the generic feature of interest, i.e., $f$ can either correspond to gravity $G$ or body acceleration $B$ in the $2\times7D$ approach, or to one among $g_l, b_l, g_r$ and $b_r$ in the $4\times4D$ approach.
We assume the following definitions.
\begin{itemize}
\item $f_{s,k}\in\mathbb{R}^n$ is the data point $k$ of feature $f$ of trial $s$, defined as:
\begin{equation}
f_{s,k} = \{f_{t,k}, f_{a,s,k}\},
\label{eq:xiks}
\end{equation}
where $f_{t,k}$ stores the time information and $f_{a,s,k}$ includes the acceleration components.
The dimension $n$ depends on the modelling approach.
\item $F^f\in\mathbb{R}^{n\times O_m}$, with $O_m = S_mK_m$, is the ordered set of data points generating the feature curve $f$ for all the $S_m$ trials, defined as:
\begin{equation}
F^f = \{f_1, \ldots, f_k, \ldots, f_{O_m}\},
\label{eq:Xixi}
\end{equation}
where
\begin{equation}
f_k = \{f_{t,k}, f_{a,k}\}
\label{eq:xik}
\end{equation}
is a generic data point taken from the whole training set, i.e., by hiding the information about the trial $s$ to which it belongs.
\end{itemize}

The purpose of GMM+GMR is to build the expected version of all features of gesture $m$, i.e.:
\begin{equation}
\hat{F}^{f} = \{\hat{f}_1, \ldots, \hat{f}_k, \ldots, \hat{f}_{K_m}\}, 
\label{eq:hatXifeat}
\end{equation}
with:
\begin{equation}
\hat{f}_k = \{f_{t,k},\hat{f}_{a,k},\hat{\Sigma}_{a,k}\}, 
\label{eq:hatxir}
\end{equation}
where $\hat{f}_{a,k}$ is the conditional expectation of $f_{a,k}$ given $f_{t,k}$ and $\hat{\Sigma}_{a,k}$ is the conditional covariance of $f_{a,k}$ given $f_{t,k}$.
The model $\hat{F}^m$ of gesture $m$ is then defined as the set of the feature models $\hat{F}^{f}$.
Please note that the number of data points in the expected curve may not be necessarily the same as the number $K_m$ of data points in the trials of the training set.
The equality is imposed here for the clarity of the description.

\begin{figure}
\centering
\subfigure[Left hand]
{\includegraphics[width=12.1cm]{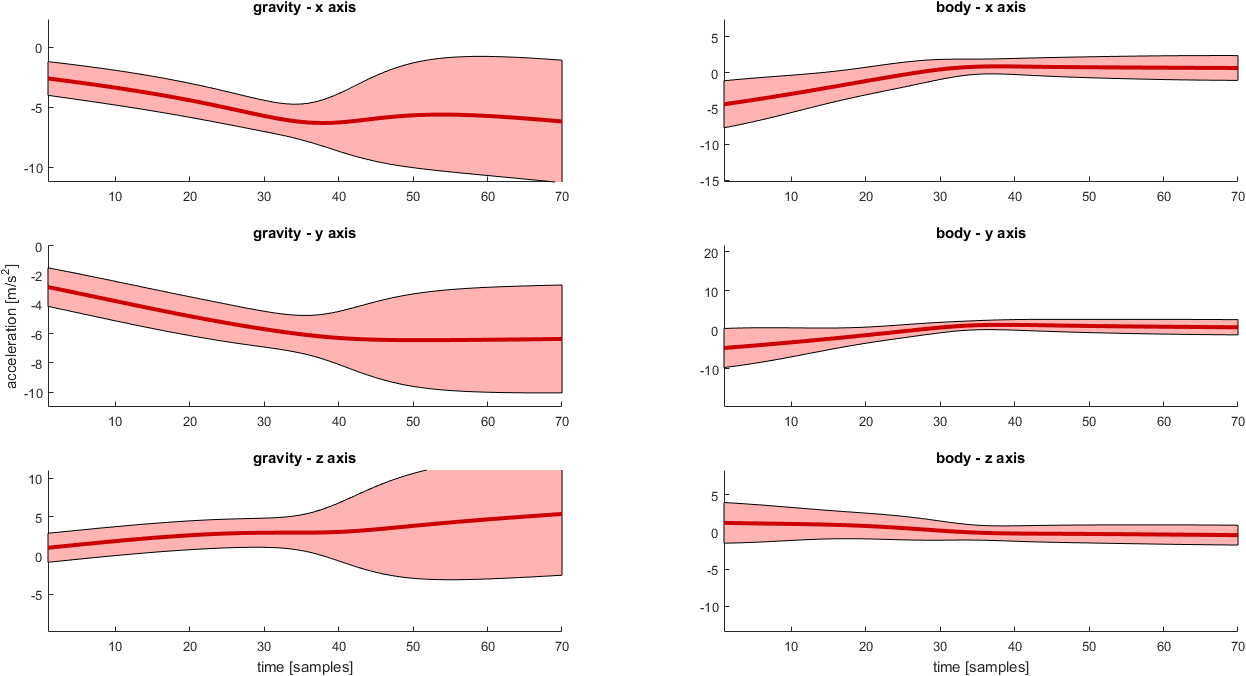}}
\subfigure[Right hand]
{\includegraphics[width=12.1cm]{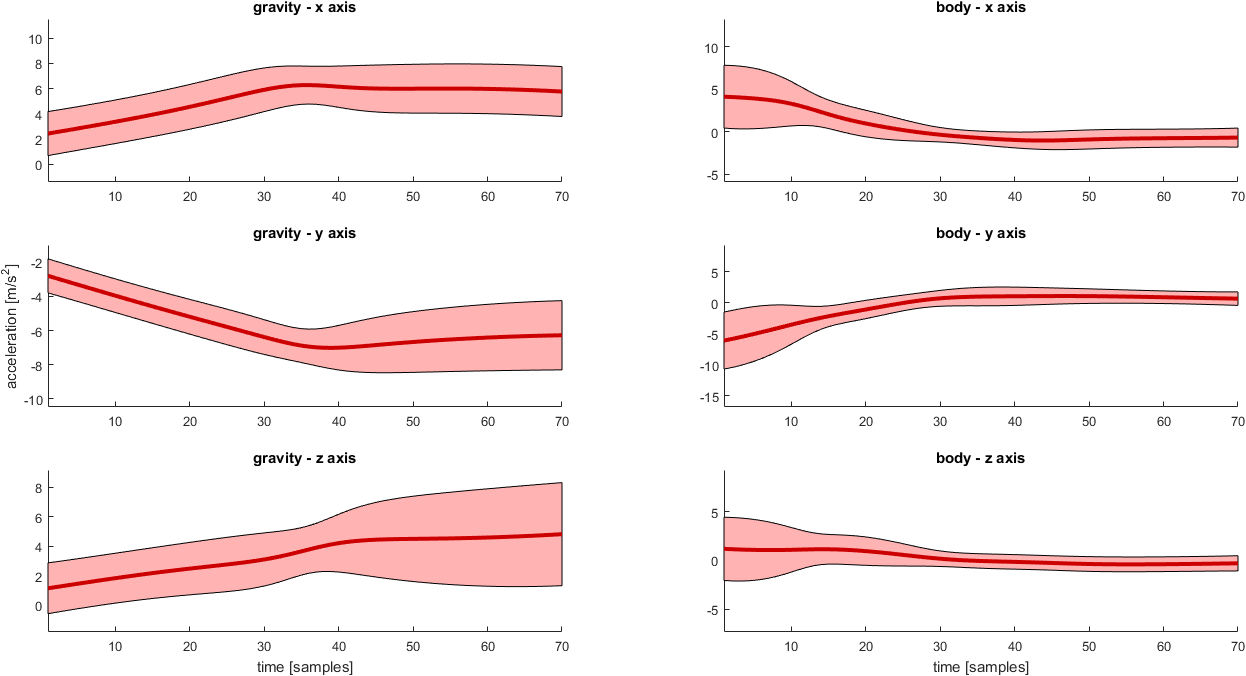}}
\caption{Model of the gesture \textit{open a wardrobe} in the implicit correlation approach ($4\times4D$), which is defined in terms of: (i) left hand gravity component (a)-left, (ii) left hand body acceleration component (a)-right, (iii) right hand gravity component (b)-left, (iv) right hand body acceleration component (b)-right.}
\label{fig:WO_4times4D}
\end{figure}

\begin{figure}
\centering
\subfigure[$l_x$, $l_y$, $l_z$ axes]
{\includegraphics[width=10.4cm]{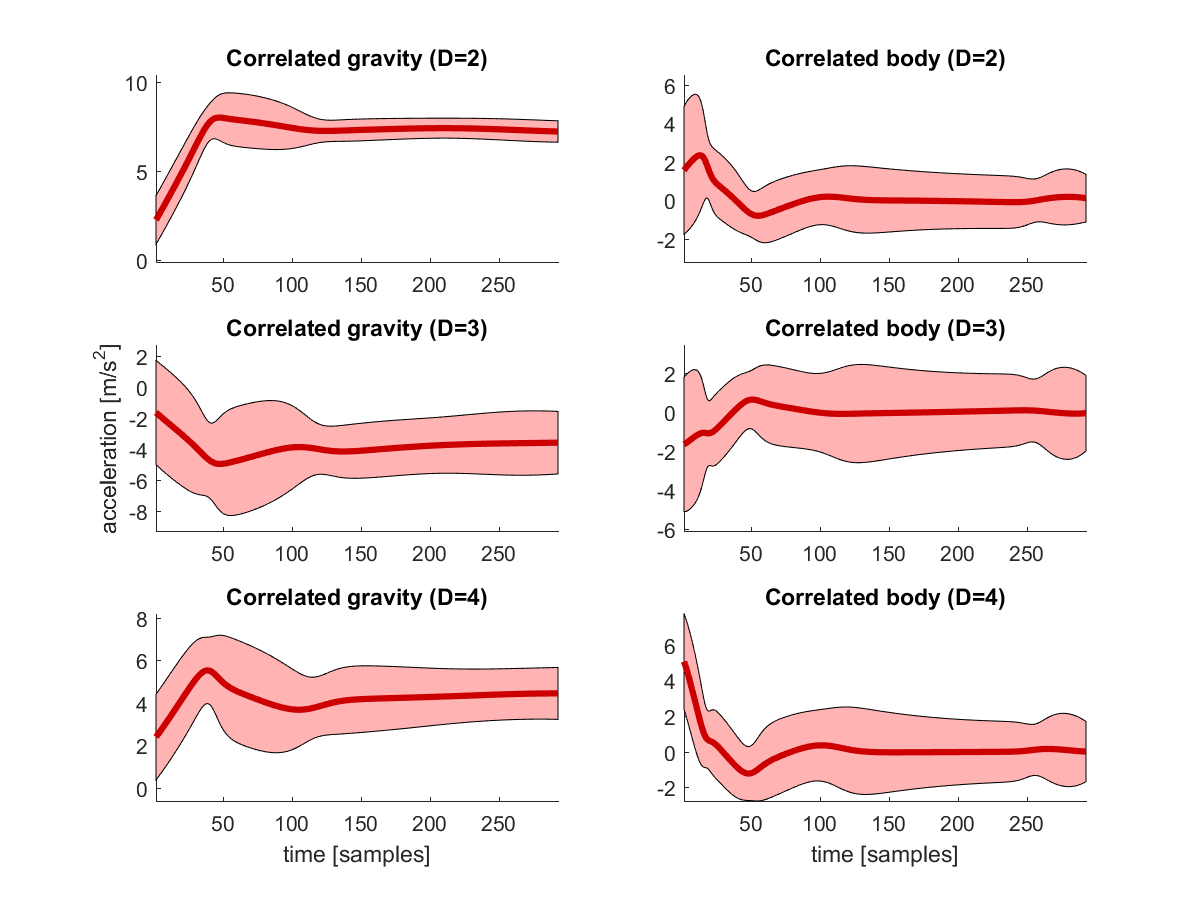}}
\subfigure[$r_x$, $r_y$, $r_z$ axes]
{\includegraphics[width=10.4cm]{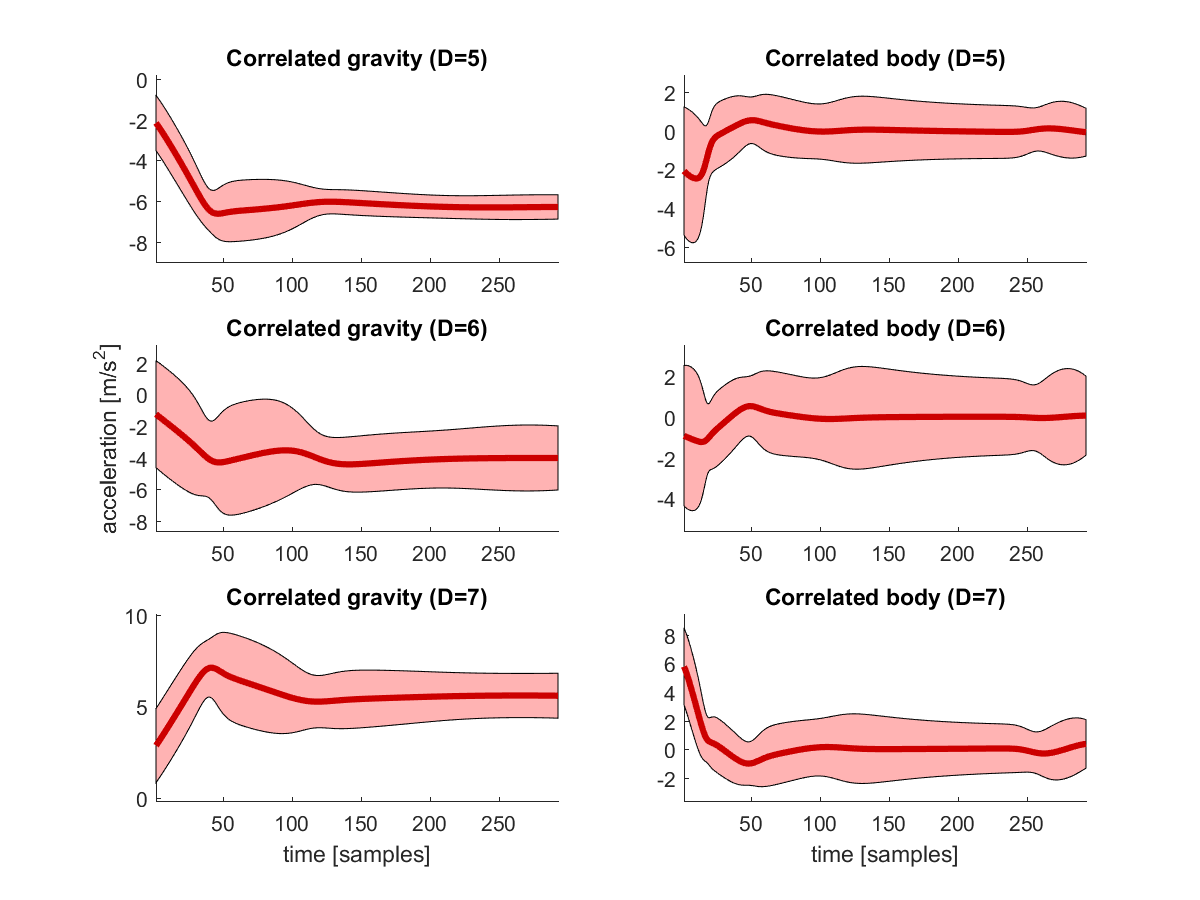}}
\caption{Model of the gesture \textit{open and close curtains} in the explicit correlation approach ($2\times7D$), which is defined in terms of: (i) gravity component (a,b)-left, (ii) body acceleration component (a,b)-right.}
\label{fig:OCC_2times7D}
\end{figure}

Figure \ref{fig:WO_4times4D} shows the $2D$ projections of the model of the gesture \textit{open a wardrobe}, computed from the full dataset of $60$ recordings with the $4\times4D$ approach.
The four modelled features are $g_l$ (a)-left, $b_l$ (a)-right, $g_r$ (b)-left, $b_r$ (b)-right.
Figure \ref{fig:OCC_2times7D} shows the $2D$ projections of the model of the gesture \textit{open and close curtains}, computed from the full dataset of $60$ recordings with the $2\times7D$ approach.
The two modelled features are $G$ (a,b)-left and $B$ (a,b)-right.
In both cases, the solid red line represents the projection of the expected curve $\hat{f}_a$ on one time-acceleration space, while the pink area surrounding it represents the conditional covariance $\hat{\Sigma}_a$.

The modelling procedure requires in input the number of Gaussian functions to use, which varies both with the gestures and with the features.
The \textit{modelling parameters estimation} stage implements a procedure based on the k-means clustering algorithm and the silhouette clustering quality metric \cite{Rousseeuw87} for the estimation of the number of Gaussian functions to use and their initialisation \cite{Bruno12}.
Other choices are equally legitimate.
We again point to the references for the details of the procedure.

\subsection{Comparison}

\begin{figure}
\centering
\includegraphics[width=11.5cm]{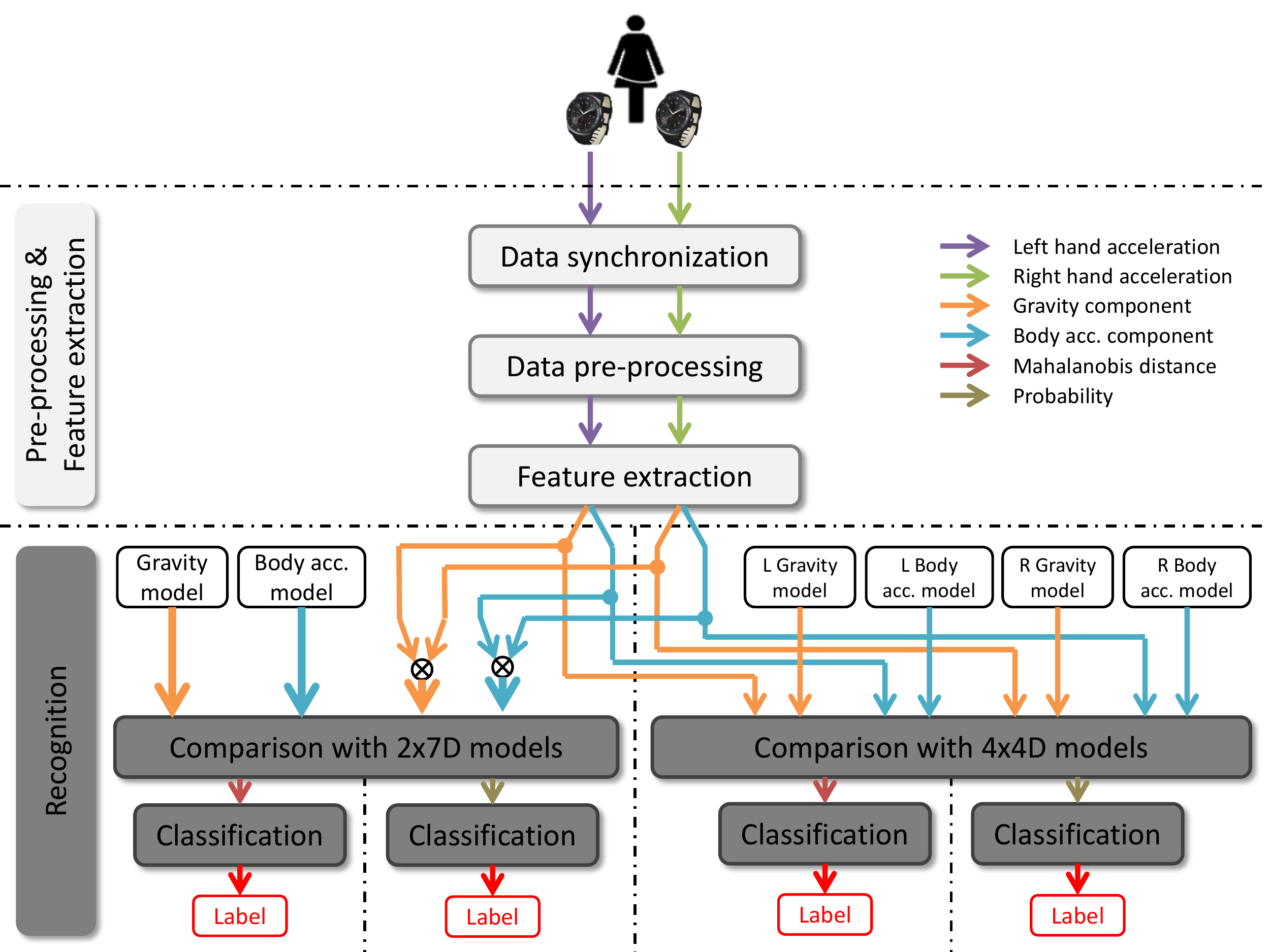}
\caption{System architecture (testing phase).}
\label{fig:testing_architecture}
\end{figure}
As it is shown in Figure \ref{fig:testing_architecture}, the testing phase executes the same procedures for data synchronisation, data pre-processing (noise reduction) and feature extraction of the training phase.
Then, in accordance with the chosen modelling approach, the features are either expressed in the form of \eqref{eq:4Dfeatures} or \eqref{eq:7Dfeatures}.

The recognition procedure is composed of a comparison stage, devoted to ranking the similarity between the testing data and each previously learned model, and a classification stage, responsible for the final labelling of the testing data on the basis of comparison results. 

We propose two \textit{comparison} procedures.
The \textit{distance}-based comparison computes the Mahalanobis distance between the testing features and each model features, while the \textit{probability}-based comparison computes the likelihood of the occurrence of the testing features given each model features.
In both cases, the \textit{classification} stage identifies the model with most prominent distance/probability, if any, and labels the testing data accordingly.
Both comparison techniques are applied for the two considered modelling approaches, thus yielding four combinations of training and testing procedures.
The comparison of their performance is the topic of the experimental evaluation presented in Section \ref{sec:experimental_evaluation}.

Let us consider a moving horizon window of length $K_w$ on the testing streams and let us denote with $F^w$ the set of features $F^{w,f}$ extracted from the window in accordance with the chosen modelling approach.

In the distance-based approach, on the basis of the $M$ available models, we compute $M$ distances $d(\hat{F}^m,F^w)$.
The Mahalanobis distance is a probabilistic distance measure used to compute the similarity between sets of random variables whose means and variances are known \cite{Mahalanobis36}.
The Mahalanobis distance $d$ between a generic element $\hat{f}_k \in \hat{F}^{f}$ defined by \eqref{eq:hatxir} and a generic element $f_{k} \in F^{w,f}$ defined in accordance with \eqref{eq:xik}, is computed as:
\begin{equation}
d(\hat{f}_k,f_{k}) = \sqrt{(\hat{f}_{a,k}-f_{a,k})^T (\hat{\Sigma}_{a,k})^{-1}(\hat{f}_{a,k}-f_{a,k})}.
\label{eq:dxiMcj}
\end{equation}
The accumulated distance between $\hat{F}^f$ and $F^{w,f}$ is computed as:
\begin{equation}
d(\hat{F}^{f},F^{w,f}) = {\sum_{k=1}^{K_m}}d(\hat{f}_k,f_{k}),
\label{eq:dphixy}
\end{equation}
that is, by integrating the distance between each element in the run-time stream and its corresponding element in the model.
The overall distance $d(\hat{F}^m,F^w)$ is computed as a weighted sum of all feature distances.
In our experiments we choose the weights to be equal for all features.
This approach weights more the precision associated with gesture production, and emphasises dexterity and robustness of bimanual motions.

In the probability-based approach, the probability of the window feature $F^{w,f}$ to be an occurrence of the model $\hat{F}^{f}$ is computed as:
\begin{equation}
\begin{split}
p(\hat{F}^{f},F^{w,f})&=\mathcal{N}(f_k,\hat{f}_{a,k},\hat{\Sigma}_{a,k}) \quad\quad\quad \forall k \in 1...K_m\\
&=\frac{1}{\sqrt{2\pi^n|\hat{\Sigma}_{a,k}|}}e^{-\frac{1}{2}(f_{a,k}-\hat{f}_{a,k})^T(\hat{\Sigma}_{a,k})^{-1}(f_{a,k}-\hat{f}_{a,k})}.
\end{split}
\label{eq:probability}
\end{equation}

The overall probability $p(\hat{F}^m,F^w)$ is computed as a weighted sum of all feature probabilities, i.e., a \textit{mixture}.
We again choose the weights to be equal for all features.
When we use probabilities, we consider the gesture as a whole, and therefore we account for small variations in the gesture execution speed.

\section{Experimental Evaluation}
\label{sec:experimental_evaluation}

\subsection{Experimental Setup}

In all the experiments, we adopt two smartwatches LG G watch R W110 as sensing devices (Android Wear 1.0, CPU Quad-Core 1.2GHz Qualcomm Snapdragon 400, 4GB/512MB RAM) equipped with a tri-axial accelerometer.
The sampling frequency is 40Hz.
The smartwatches automatically sync on startup with the smartphone they are paired with.
By pairing the two smartwatches with the same smartphone, we ensure that they are synchronised with each other with a precision satisfying the requirements of our application.

\begin{figure}
\centering
{\includegraphics[width=12cm]{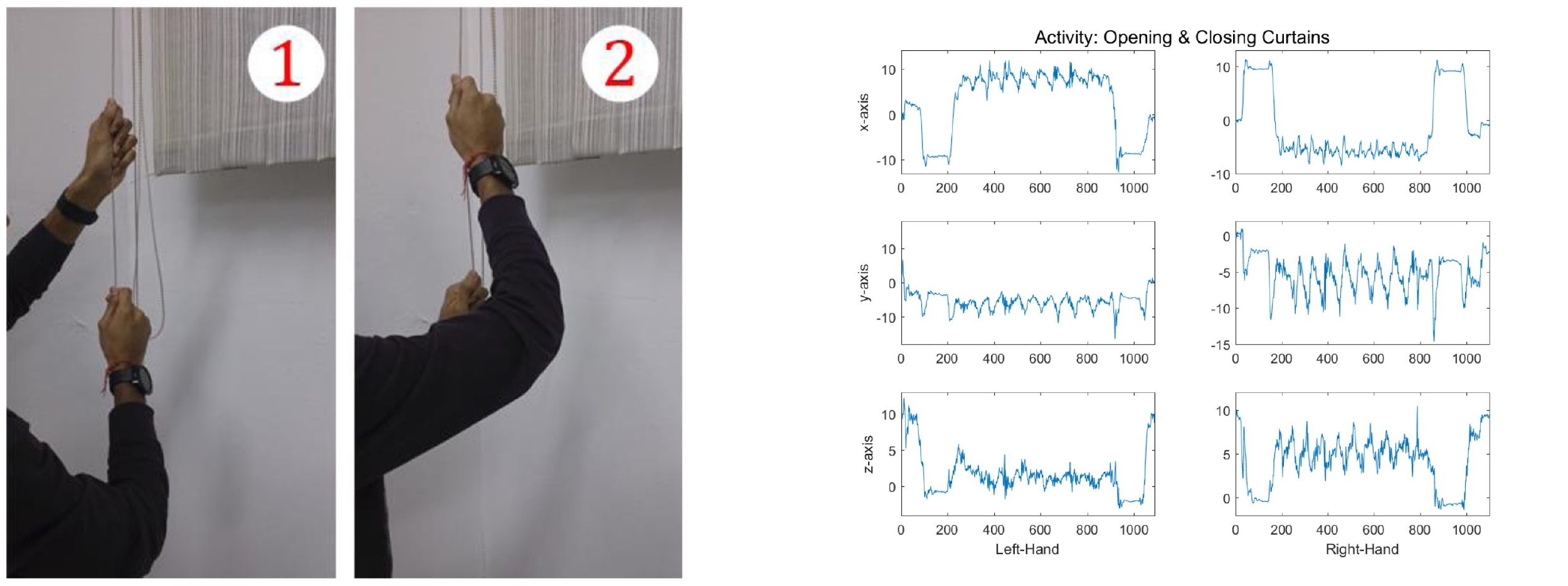}}
\caption{Open and close curtains.}
\label{fig:OCC}
\end{figure}

\begin{figure}
\centering
{\includegraphics[width=12cm]{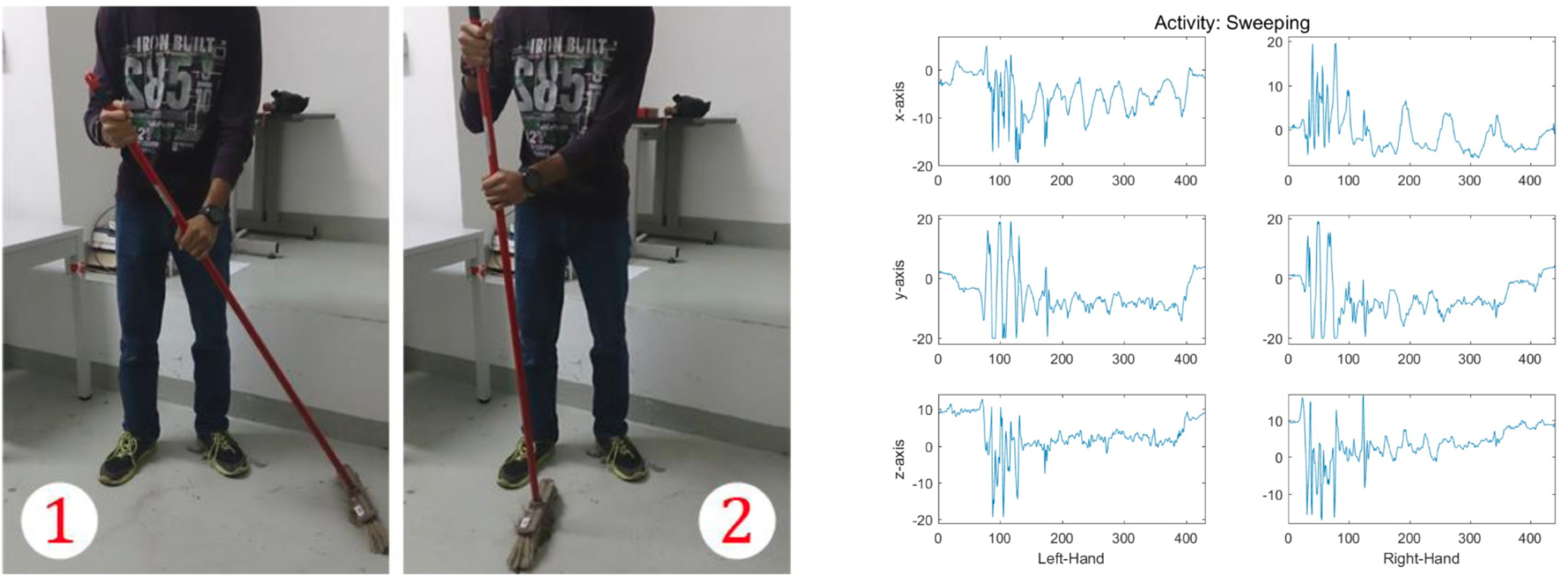}}
\caption{Sweep the floor.}
\label{fig:Swp}
\end{figure}

\begin{figure}
\centering
{\includegraphics[width=12cm]{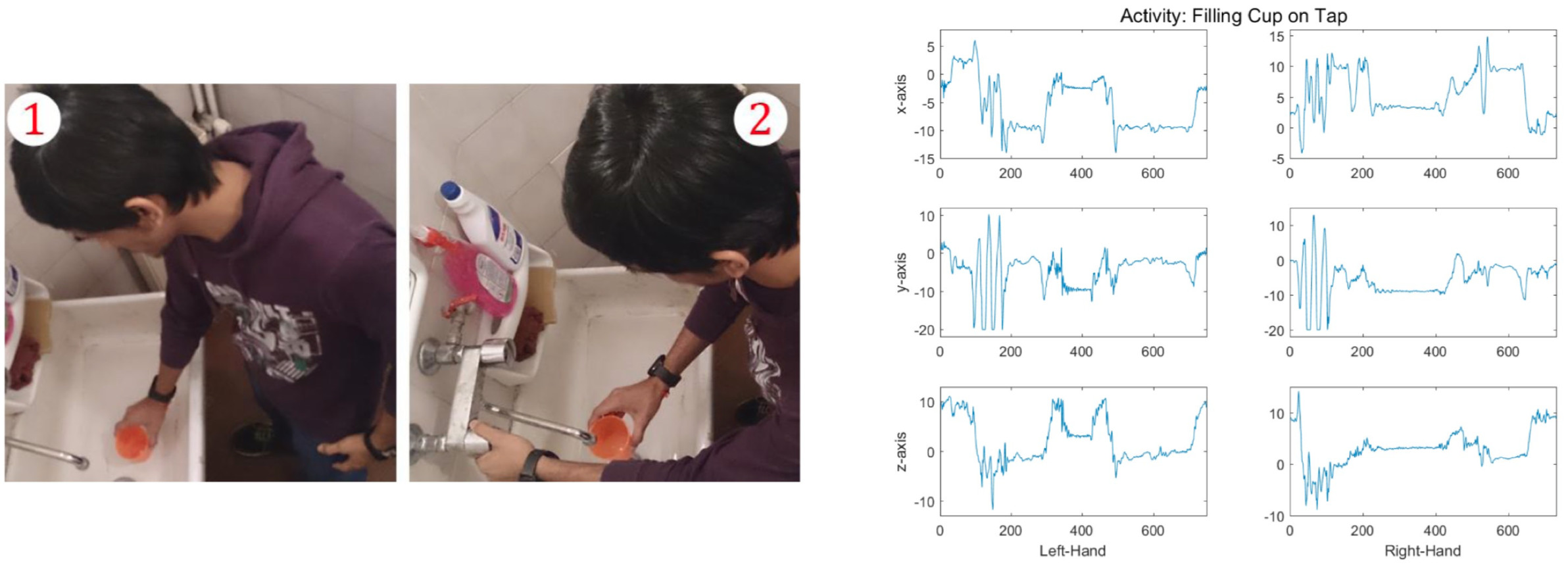}}
\caption{Fill a cup with tap water.}
\label{fig:FCoT}
\end{figure}

\begin{figure}
\centering
{\includegraphics[width=12cm]{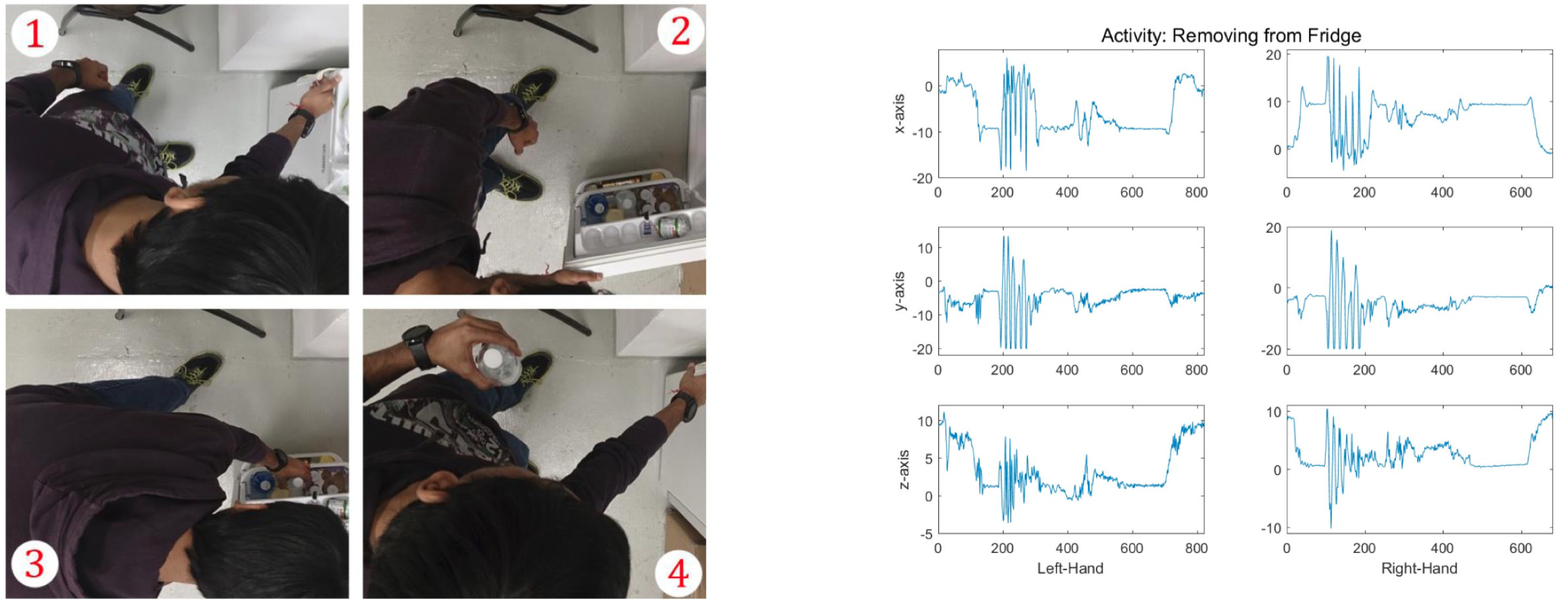}}
\caption{Take a bottle from the fridge.}
\label{fig:RfF}
\end{figure}

\begin{figure}
\centering
{\includegraphics[width=12cm]{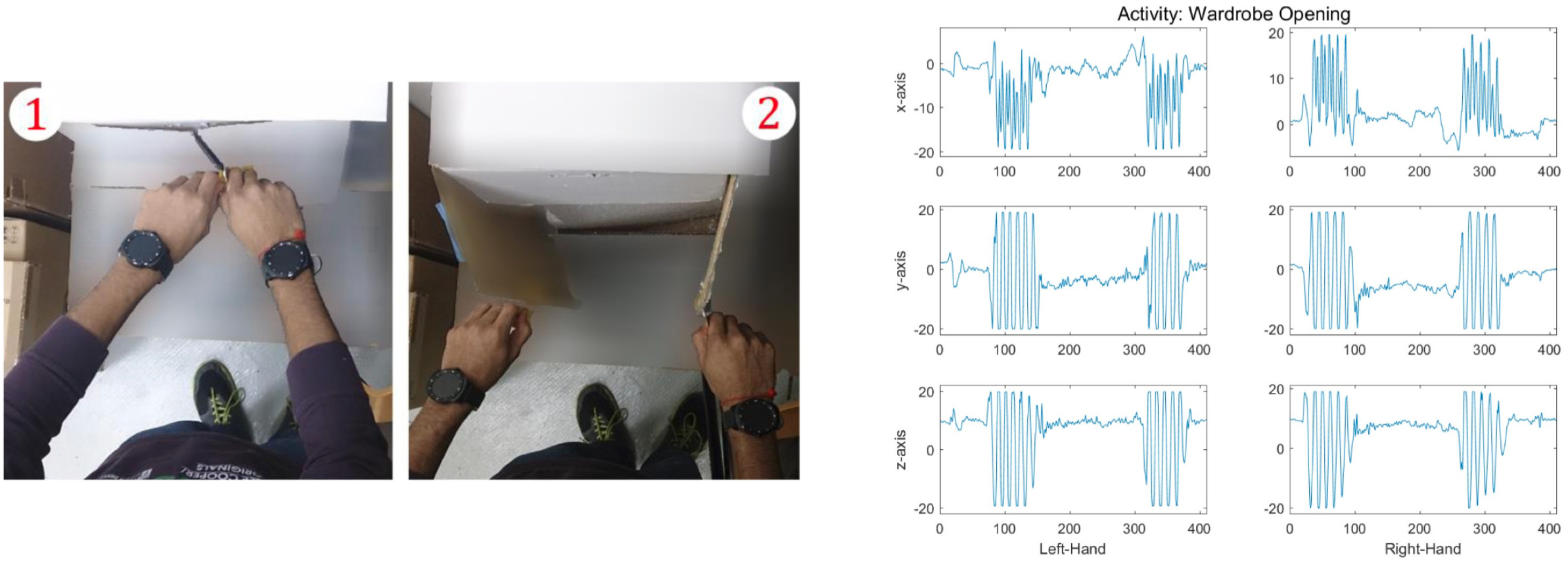}}
\caption{Open a wardrobe.}
\label{fig:WO}
\end{figure}

We consider five bimanual gestures:
\begin{itemize}
\item \textit{Open and close curtains} (OCC).
Extend and retract lateral-sliding curtains by pulling the connecting cords with an alternated up-and-down movement of the hands.
Keep pulling until the curtain is fully closed or opened (see Figure \ref{fig:OCC}).
\item \textit{Sweep the floor} (SWP).
Pull a conventional broom from left to right, to sweep the floor.
Three strokes are required (see Figure \ref{fig:Swp}).
\item \textit{Fill a cup with tap water} (FCOT).
With the right hand, take a cup from the sink and hold it below the tap, while rotating the tap's handle with the left hand to fill the cup with water (see Figure \ref{fig:FCoT}).
\item \textit{Take a bottle from the fridge} (RFF).
With the right hand, open the door of a small fridge, then, with the left one, take a bottle from it.
Lastly, close the fridge door with the right hand (see Figure \ref{fig:RfF}).
\item \textit{Open a wardrobe} (WO).
Open a two-doors small wardrobe moving the two hands concurrently (see Figure \ref{fig:WO}).
\end{itemize}

Intuitively, the gestures \textit{open and close curtains}, \textit{sweep the floor} and \textit{open a wardrobe} fully constrain the movement of the two hands, while the gestures \textit{fill a cup with tap water} and \textit{take a bottle from the fridge} allow for more freedom in their execution, as far as synchronisation between the arms is concerned.
Moreover, the gestures \textit{sweep the floor} and \textit{open a wardrobe} imply that the two hands are moved concurrently, while the gestures \textit{open and close curtains}, \textit{fill a cup with tap water} and \textit{take a bottle from the fridge} mostly require the two hands to be moved in sequence.
Lastly, the gestures \textit{open and close curtains} and \textit{sweep the floor} are recursive, i.e., composed of a number of repetitions of simpler movements, while the gestures \textit{open a wardrobe}, \textit{fill a cup with tap water} and \textit{take a bottle from the fridge} are non-recursive.

Figures \ref{fig:OCC}-\ref{fig:WO} show pictures of an execution of the gesture on the left hand side and the acceleration measured at the two wrists during the execution of the same gesture on the right hand side.
The impact of the aforementioned gesture characteristics (constrained/not constrained, concurrent/sequential, recursive/not recursive) on the accelerations measured at the wrist, and therefore on the considered modelling and comparison procedures, is not known: finding it is one of the goals of the experiments we have conducted.

For each gesture, we collected a dataset of $60$ recordings from $10$ volunteers with age ranging from $22$ to $30$ years old.
The volunteers, wearing the smartwatches, have been asked to autonomously start and stop the recordings and, once an experimenter described the gesture, to perform it in a natural way.
All repetitions were supervised.
In addition to this dataset, we asked a number of volunteers to take some recordings in real-life conditions.
They have been asked to clean a room and, amidst the other activities, to perform the five bimanual gestures of interest.
Volunteers could freely choose the timing and sequence of the gestures, and the choices have been annotated by an experimenter.

\subsection{Performance Analysis}

We tested the four combinations of modelling and comparison procedures in terms of the standard statistical measures of accuracy, precision and recall by using k-fold cross validation on the collected dataset.
For all gestures, we split the dataset in $6$ groups of $10$ recordings each and iteratively used $5$ groups as training dataset and the remaining group for validation.

\begin{figure}
\centering
{\includegraphics[width=11cm]{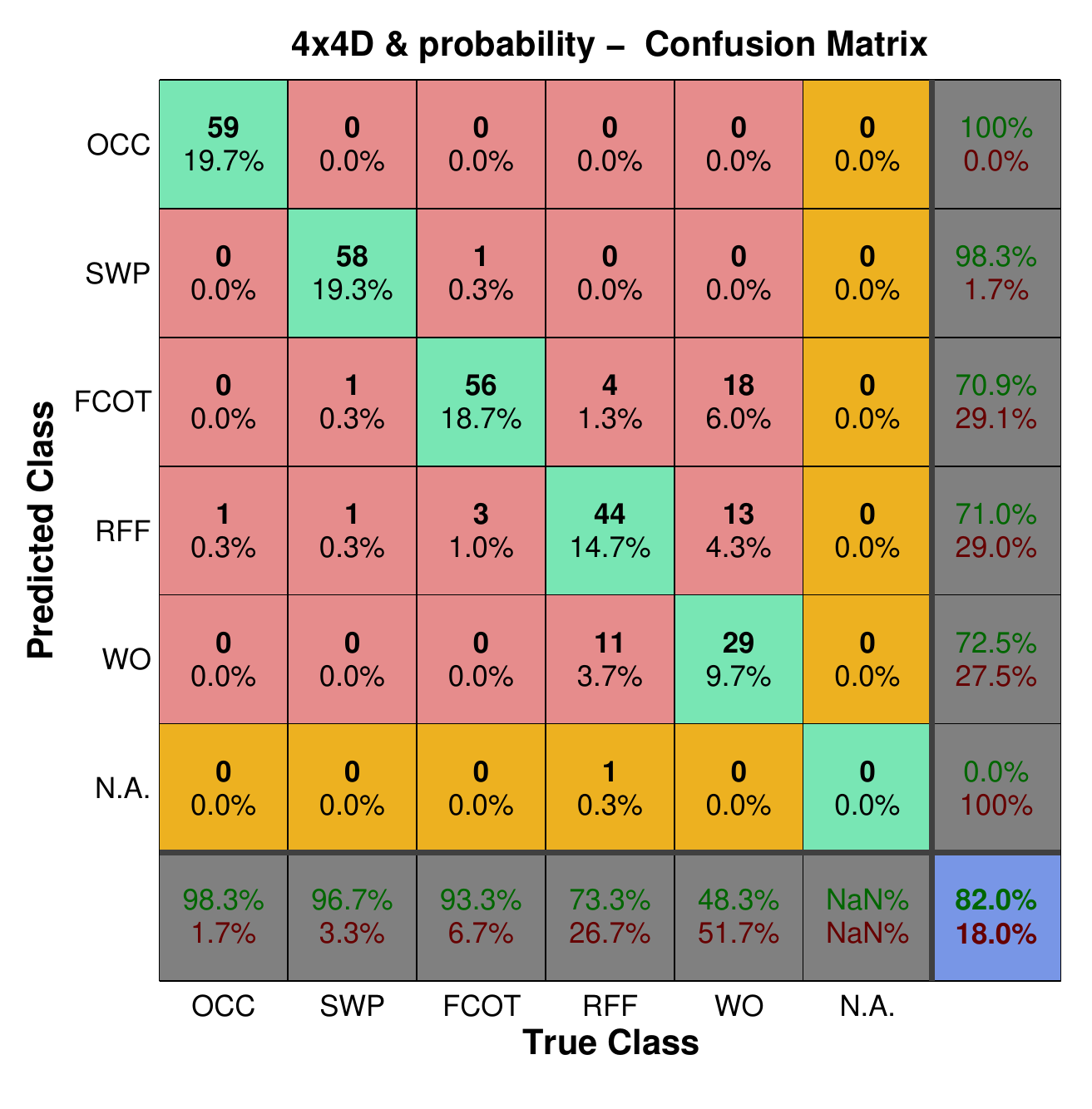}}
\caption{Results of k-fold cross validation for the $4\times4D$ modelling approach and probability-based comparison. The bottom row reports the \textit{recall} measures, while the rightmost column reports the \textit{precision} measures. The purple cell reports the overall \textit{accuracy} of the system.}
\label{fig:validation_4Dprob}
\end{figure}

\begin{figure}
\centering
{\includegraphics[width=11cm]{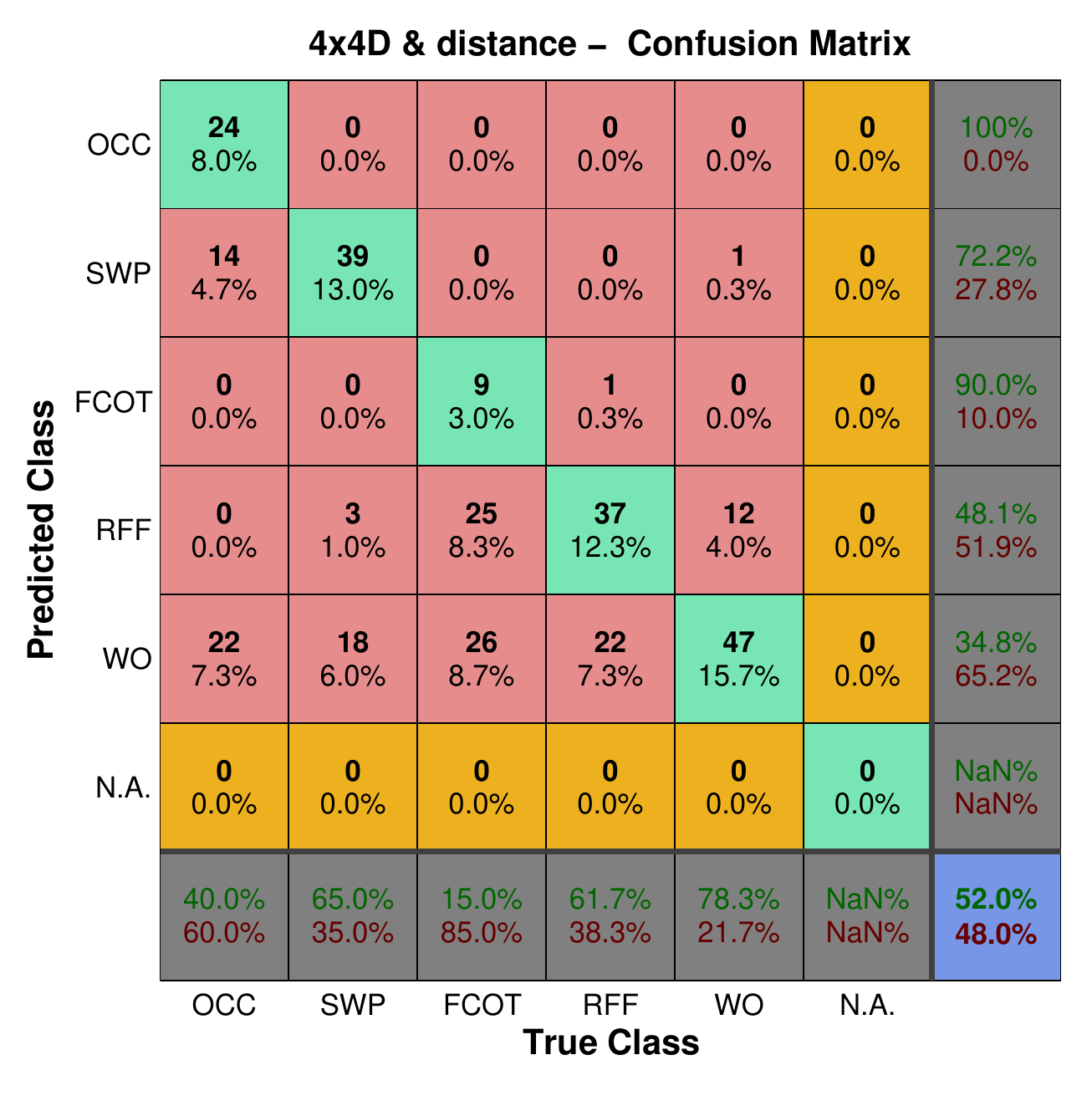}}
\caption{Results of k-fold cross validation for the $4\times4D$ modelling approach and distance-based comparison. The bottom row reports the \textit{recall} measures, while the rightmost column reports the \textit{precision} measures. The purple cell reports the overall \textit{accuracy} of the system.}
\label{fig:validation_4Ddist}
\end{figure}

\begin{figure}
\centering
{\includegraphics[width=11cm]{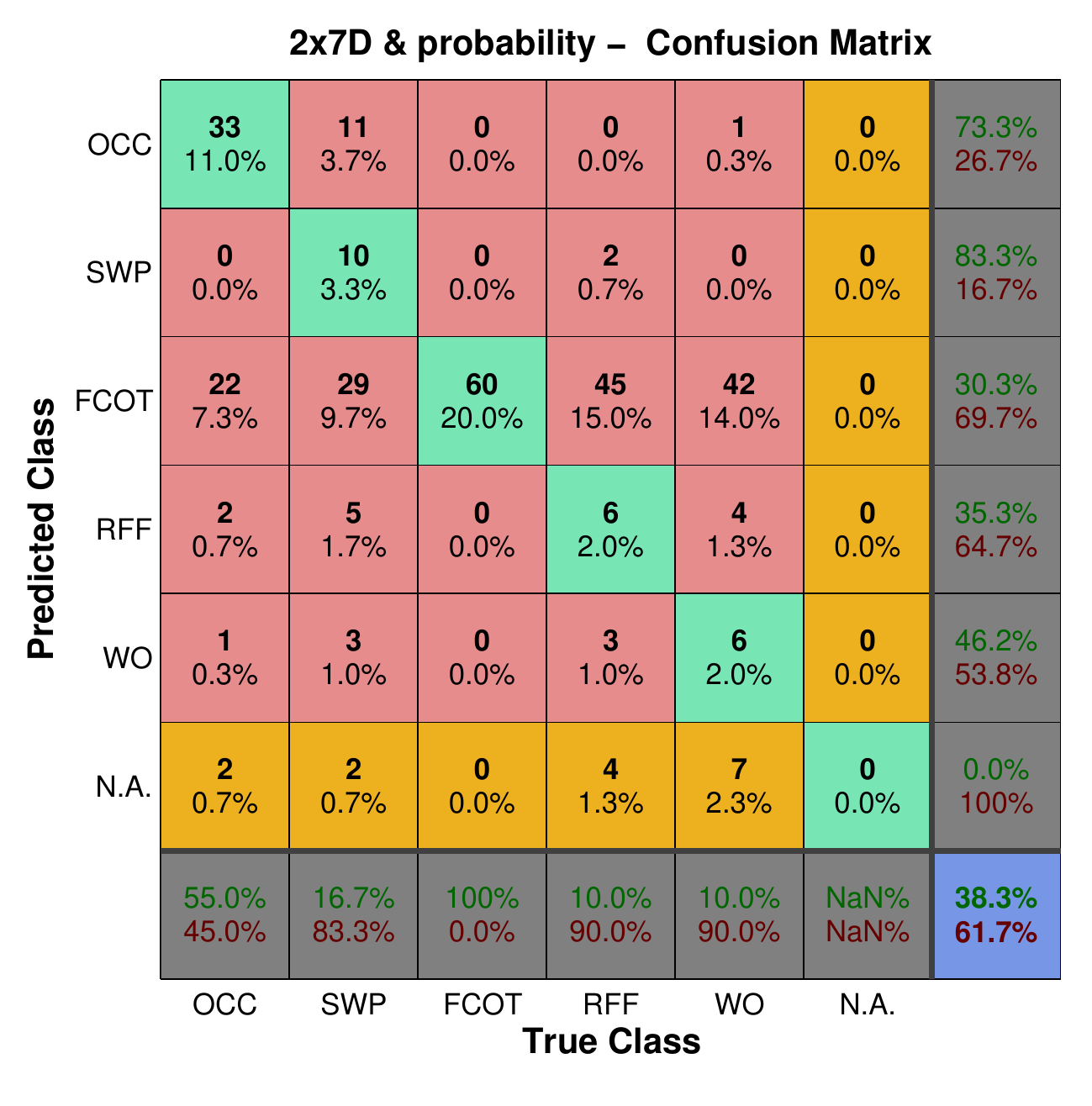}}
\caption{Results of k-fold cross validation for the $2\times7D$ modelling approach and probability-based comparison. The bottom row reports the \textit{recall} measures, while the rightmost column reports the \textit{precision} measures. The purple cell reports the overall \textit{accuracy} of the system.}
\label{fig:validation_7Dprob}
\end{figure}

\begin{figure}
\centering
{\includegraphics[width=11cm]{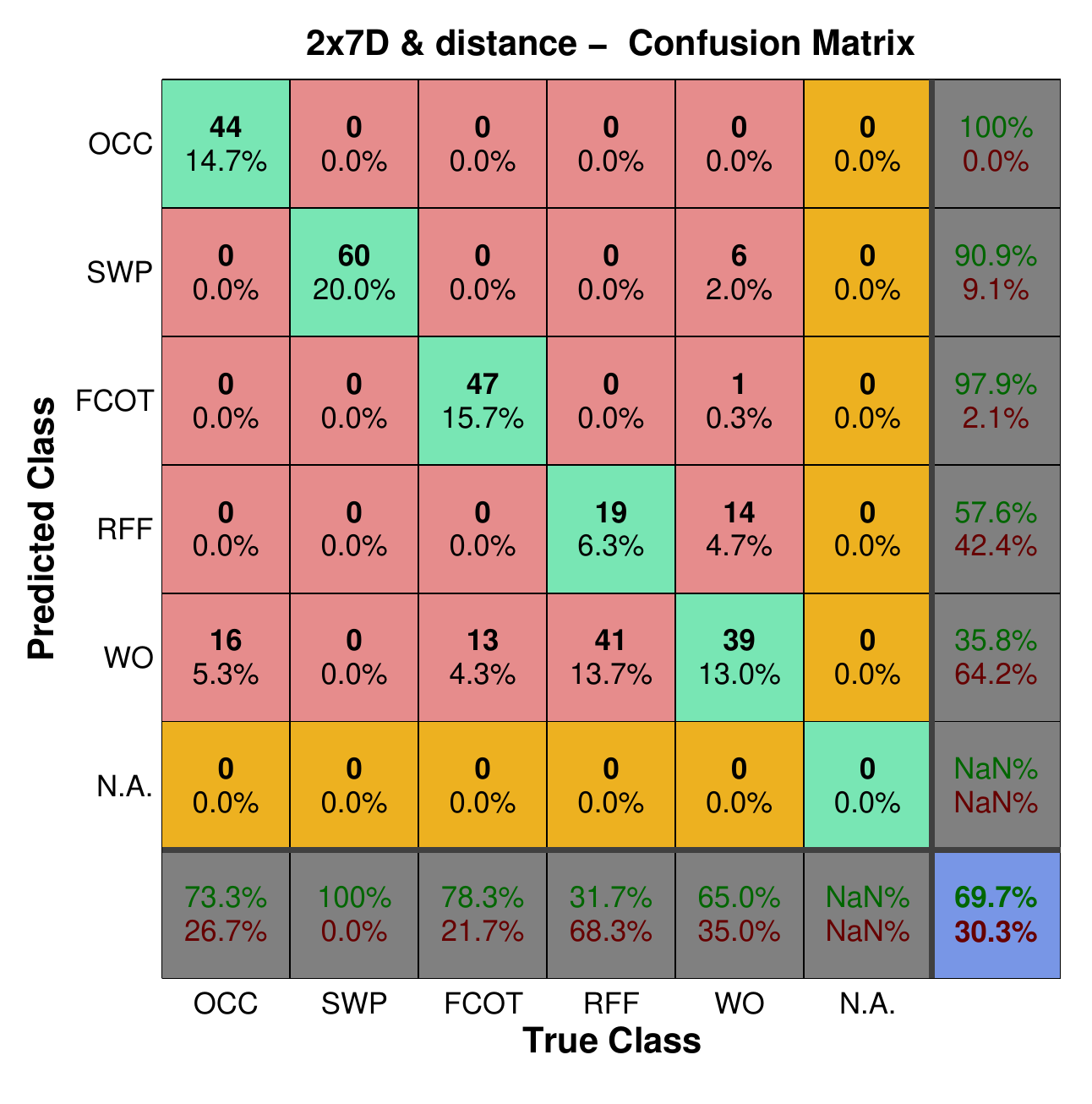}}
\caption{Results of k-fold cross validation for the $2\times7D$ modelling approach and distance-based comparison. The bottom row reports the \textit{recall} measures, while the rightmost column reports the \textit{precision} measures. The purple cell reports the overall \textit{accuracy} of the system.}
\label{fig:validation_7Ddist}
\end{figure}

The results of the k-fold cross validation on the four combinations of modelling and comparison procedures are shown in Figures \ref{fig:validation_4Dprob}-\ref{fig:validation_7Ddist}.
In all figures, the first five rows/columns refer to the gestures \textit{open and close curtains} (OCC), \textit{sweep the floor} (SWP), \textit{fill a cup with tap water} (FCOT), \textit{take a bottle from the fridge} (RFF) and \textit{open a wardrobe} (WO).
The yellow row/column collectively represents all the gestures which are not among the modelled ones (N.A.).
The columns denote the true labels of all validation recordings (i.e., the gestures actually performed during each of them), while the rows denote the labels assigned by the recognition system.
Since, collectively, the validation dataset is composed of $60$ recordings per modelled gesture, a perfect recognition system would show $60$ recordings in the first five green cells (i.e., with the predicted label matching the true label) and none in the red cells (meaning that the recording of a gesture has been classified as an occurrence of another gesture) or in the yellow cells (meaning that the recording of a gesture has been classified as an occurrence of an unknown gesture).

Column $1$ of Figure \ref{fig:validation_4Dprob} reports that, out of $60$ validation recordings actually referring to the gesture \textit{open and close curtains}, $59$ were correctly labelled as occurrences of that gesture and one was labelled as an occurrence of the gesture \textit{take a bottle from the fridge}.
This analysis allows for computing the \textit{recall} performance of the system, which corresponds to the ratio between the number of recordings of gesture $m$ correctly labelled as occurrences of gesture $m$ and the overall number of recordings of gesture $m$.
The recall values for all gestures are listed in the bottom row of the confusion matrix. 

A similar analysis of the rows of the confusion matrix allows for assessing the \textit{precision} performance of the system.
As an example, row $2$ of Figure \ref{fig:validation_4Dprob} reports that, out of $59$ recordings labelled as occurrences of the gesture \textit{sweep the floor}, $58$ were true recordings of that gesture, while one was a recording of the gesture \textit{fill a cup with tap water}.
The precision metric measures the ratio between number of recordings of gesture $m$ correctly labelled as occurrences of gesture $m$ and the overall number of recordings labelled as occurrences of gesture $m$. The precision values for all gestures are listed in the rightmost column of the confusion matrix.

Lastly, the aggregated analysis of the number of correct labels over the total number of recordings, i.e., the \textit{accuracy} performance of the system, is reported in the purple cell at the bottom-right corner.
As an example, the recognition system adopting the implicit correlation modelling approach ($4\times4D)$ and probability-based comparison, whose confusion matrix in shown in Figure \ref{fig:validation_4Dprob}, has an overall accuracy of $82\%$.

\subsection{Real-life Conditions}

\begin{table}
\centering
\caption{Experiment scripts for the real-life tests.}
\label{tab:real-life}
\begin{tabular}{@{ }c@{ } @{ }c@{ }}
\hline
Scenario \#1 & Scenario \#2 \\
\hline 
\multirow{2}{*}{SWP} & WO \\ 
 & RFF \\ 
\multirow{2}{*}{SWP} & FCOT \\ 
 & OCC \\ 
\hline
\end{tabular} 
\end{table}

As an additional test for assessing the system's performance, we asked two volunteers to wear the smartwatches at the wrists while carrying out cleaning chores of their choice in a room.
The first volunteer was instructed to perform twice, amidst the other activities, the gesture \textit{sweep the floor}, while the second volunteer was instructed to perform each of the other gestures once, as summarised in Table \ref{tab:real-life}.
Both tests were supervised by an experimenter, annotating the time when each gesture of interest was performed.

\begin{figure}
\centering
\includegraphics[width=12cm]{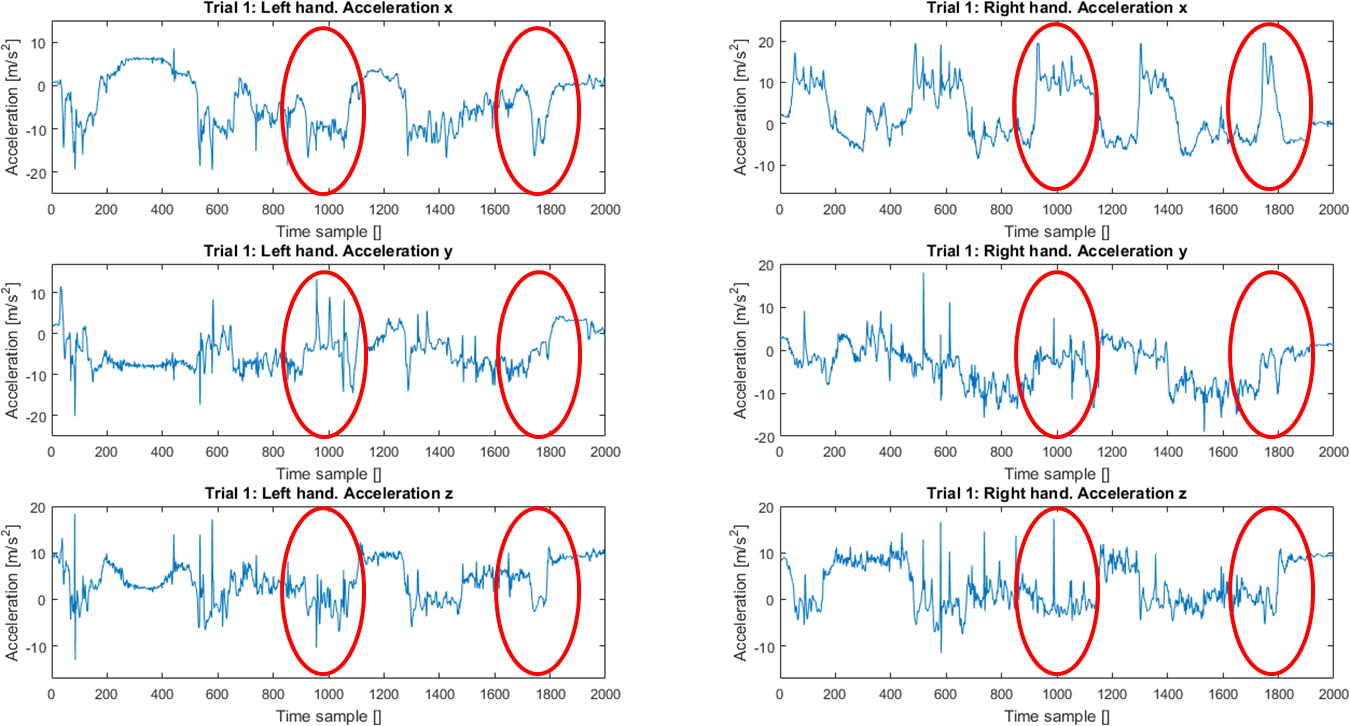}
\caption{Acceleration streams registered during the execution of the real-life scenario 1. Red circles denote the two executions of the gesture \textit{sweep the floor}.}
\label{fig:scenario1}
\end{figure}

\begin{figure}
\centering
\includegraphics[width=12cm]{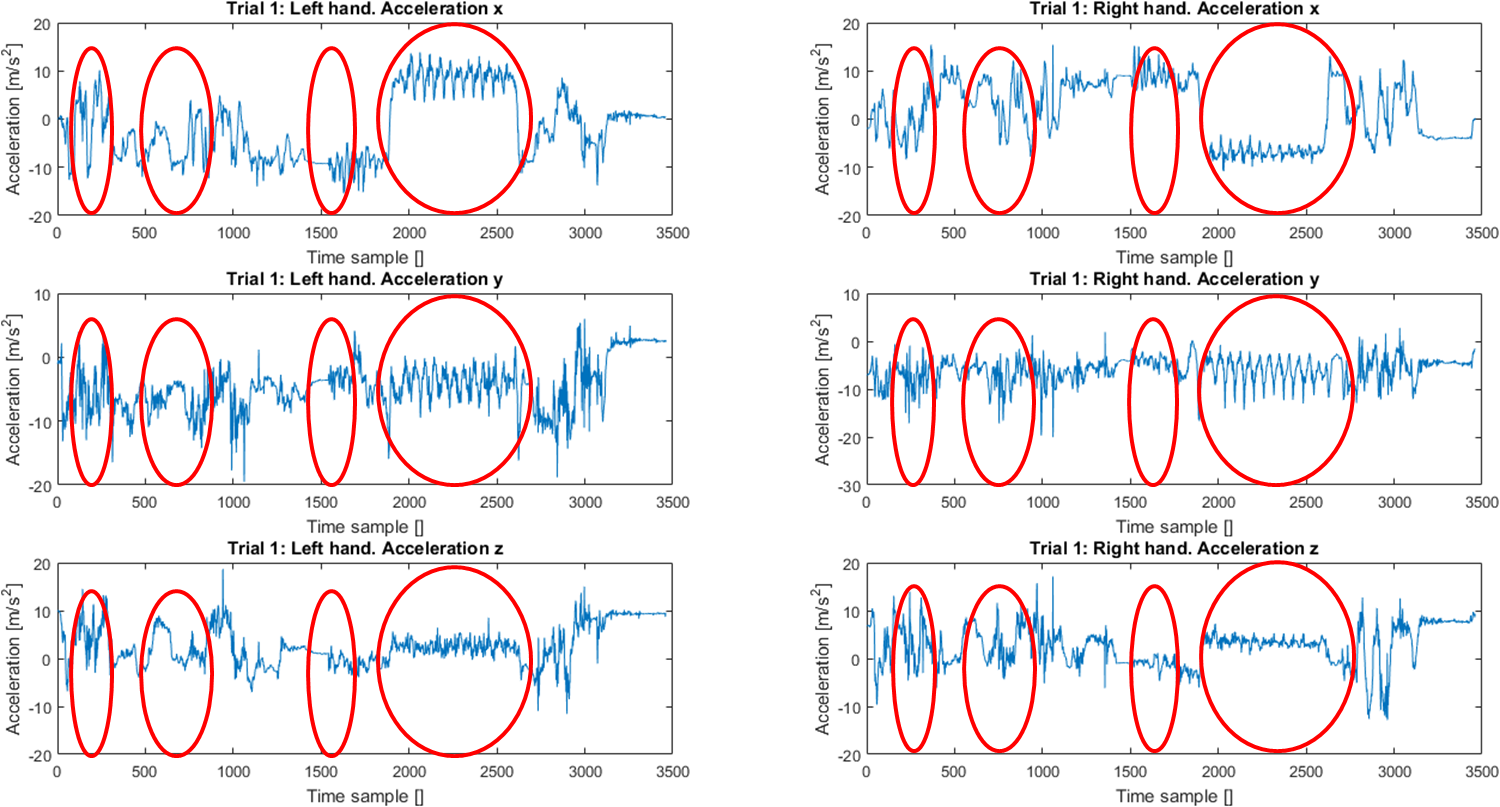}
\caption{Acceleration streams registered during the execution of the real-life scenario 2. Red circles denote, from left to right, the execution of the gestures \textit{open a wardrobe}, \textit{take a bottle from the fridge}, \textit{fill a cup with tap water} and \textit{open and close curtains}.}
\label{fig:scenario2}
\end{figure}

Figure \ref{fig:scenario1} shows the accelerations registered at the wrists of the first volunteer, while performing the real-life scenario 1, while Figure \ref{fig:scenario2} shows the accelerations registered at the wrists of the second volunteer, while performing the real-life scenario 2.
The red circles denote all occurrences of the modelled bimanual gestures.
For these tests we used the whole dataset of $60$ recordings per gesture previously described as training set for the creation of the models.

\begin{figure}
\centering
\includegraphics[width=12cm]{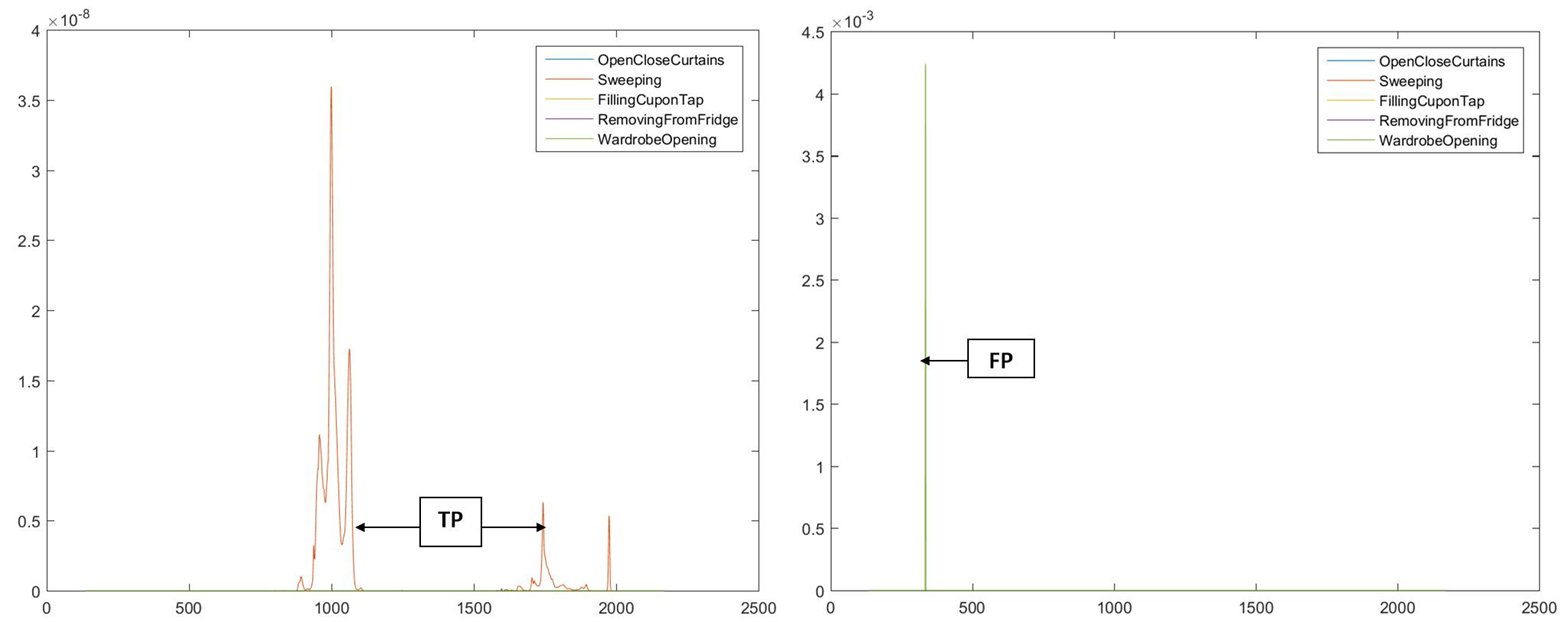}
\caption{Output of the system for the recording of scenario 1 shown in Figure \ref{fig:scenario1}. The left hand graph refers to the $4\times4D$ modelling approach and probability-based comparison, while the right hand side refers to the $4\times4D$ modelling approach and distance-based comparison.}
\label{fig:validation_scenario1_4D}
\end{figure}

\begin{figure}
\centering
\includegraphics[width=12cm]{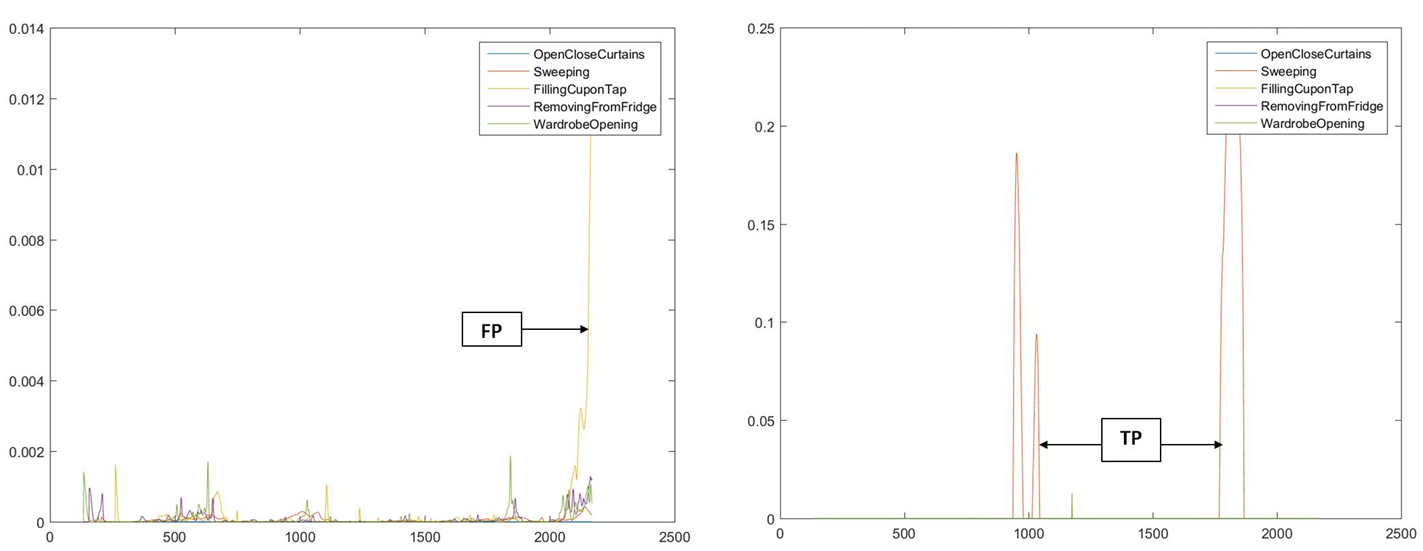}
\caption{Output of the system for the recording of scenario 1 shown in Figure \ref{fig:scenario1}. The left hand graph refers to the $2\times7D$ modelling approach and probability-based comparison, while the right hand side refers to the $2\times7D$ modelling approach and distance-based comparison.}
\label{fig:validation_scenario1_7D}
\end{figure}

\begin{figure}
\centering
\includegraphics[width=12cm]{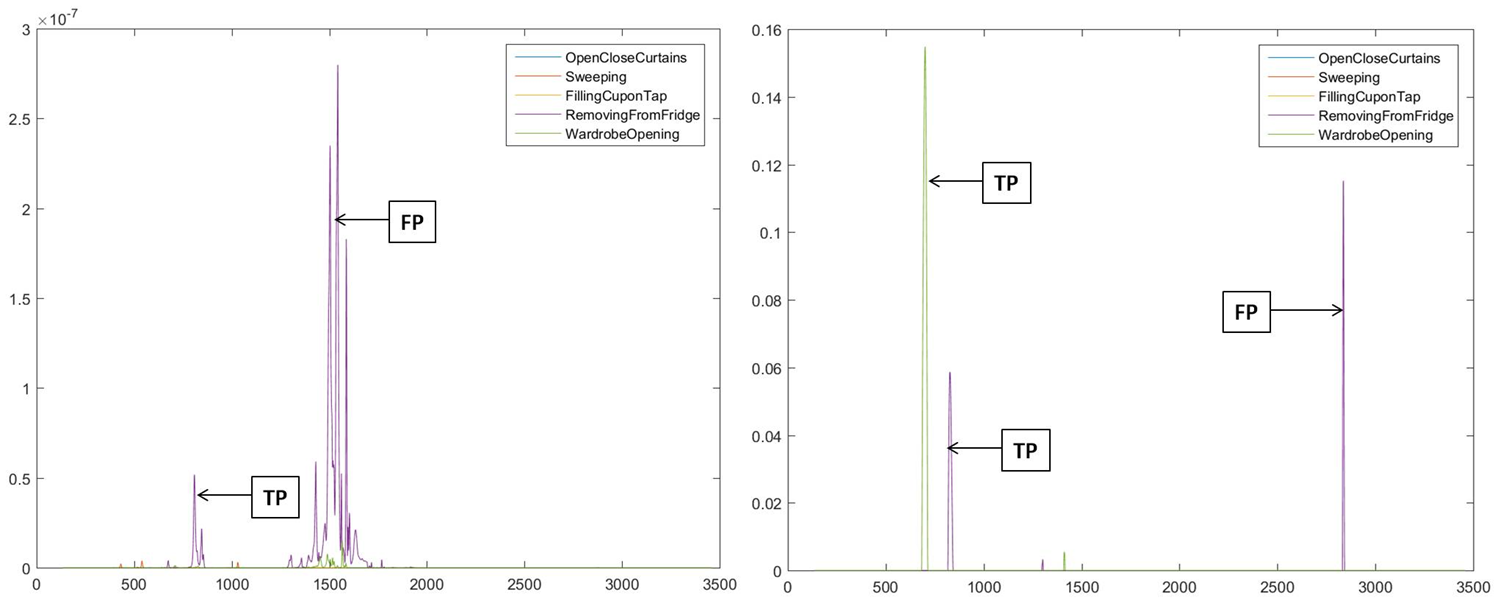}
\caption{Output of the system for the recording of scenario 2 shown in Figure \ref{fig:scenario2}. The left hand graph refers to the $4\times4D$ modelling approach and probability-based comparison, while the right hand side refers to the $4\times4D$ modelling approach and distance-based comparison.}
\label{fig:validation_scenario2_4D}
\end{figure}

\begin{figure}
\centering
\includegraphics[width=12cm]{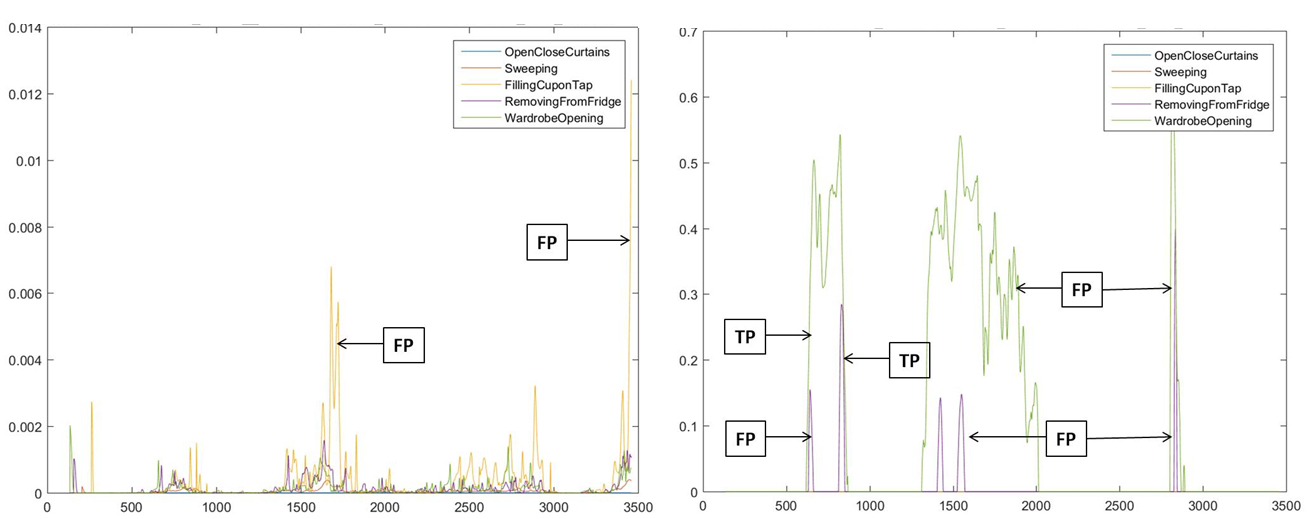}
\caption{Output of the system for the recording of scenario 2 shown in Figure \ref{fig:scenario2}. The left hand graph refers to the $2\times7D$ modelling approach and probability-based comparison, while the right hand side refers to the $2\times7D$ modelling approach and distance-based comparison.}
\label{fig:validation_scenario2_7D}
\end{figure}

Figures \ref{fig:validation_scenario1_4D}-\ref{fig:validation_scenario2_7D} show the output of the recognition system, for the first (Figures \ref{fig:validation_scenario1_4D} and \ref{fig:validation_scenario1_7D}) and the second scenario (Figures \ref{fig:validation_scenario2_4D} and \ref{fig:validation_scenario2_7D}), for the four combinations of modelling and comparison procedures.
In all figures, the x-axis denotes time and the y-axis denotes the probability or inverse distance value of the models.
The \textit{TP} box marks the true positive recognitions, i.e., all the occurrences of a modelled gesture which are correctly recognised.
The \textit{FP} box marks the false positive recognitions, i.e., all the occurrences of gestures which have been assigned the wrong label.

\subsection{Discussion}

As Figures \ref{fig:validation_4Dprob}-\ref{fig:validation_7Ddist} show, the combination allowing for the best recognition performance relies on the \textit{implicit} correlation modelling approach and the \textit{probability-based} comparison: this system achieves an overall accuracy of $82\%$ (see Figure \ref{fig:validation_4Dprob}) and it retains good recognition performance, especially in terms of precision, for all modelled gestures (the gesture with worst recall is \textit{open a wardrobe}, with $48.3\%$, while the gesture with worst precision is \textit{fill a cup with tap water}, with $70.9\%$).
Interestingly enough, the combination resulting in the worst recognition performance ($38.3\%$ overall accuracy) relies on the explicit correlation modelling approach and the same probability-based comparison (see Figure \ref{fig:validation_7Dprob}), thus confirming that there is a tight relation between the modelling and comparison procedures.
The same modelling approach, with the distance-based comparison procedure, significantly improves its performance ($69.7\%$ overall accuracy, see Figure \ref{fig:validation_7Ddist}).

For all four considered combinations, the gestures \textit{open and close curtains} and \textit{sweep the floor} consistently achieve the highest precision and recall values, which seems to suggest that recursive motions (i.e., composed of the repeated execution of simple movements), regardless whether they involve the concurrent or sequential motion of the two arms, are easier to model and recognise.
In other words, repeated gestures produce more stable patterns as far as the detection and classification system is concerned.
Conversely, the gestures \textit{open a wardrobe} and \textit{take a bottle from the fridge} consistently achieve the lowest precision and recall values, with the latter performing especially poorly with the implicit correlation modelling approach.
These results, albeit preliminary, seem to confirm our intuition that the explicit modelling approach is much more sensitive to small differences between run-time streams and the corresponding model than the implicit correlation modelling approach.

In accordance with the k-fold validation analysis, also in the real-life tests the combinations achieving the best performance are the implicit correlation modelling approach with probability-based comparison and the explicit correlation modelling approach with distance-based comparison, which, on the whole, successfully recognise respectively three and four of the six gestures of interest.

Real-life tests highlight a difference in the pattern of the recognition labels between the probability-based and distance-based comparison procedures.
As Figures \ref{fig:validation_scenario1_4D}-\ref{fig:validation_scenario2_7D} show, in the second case the labels follow a smoother pattern, which reduces the number of false positive recognition.
The adoption of reasoning techniques analysing label patterns to increase the recognition accuracy, has been proved effective in the case of a single stream \cite{Bruno14b} and may lead to significant improvements also in the case of bimanual gestures.

What these results have to say about the way we currently understand bimanual gestures in humans?
Not surprisingly, the best results we achieve involve implicit correlation with probability-based comparison.
Whilst with implicit correlation we do not assume the motion of the two hands/arms to be correlated explicitly, probability-based comparison is more robust with respect to small variations in the phases of the two hands/arms.
If we had assumed the two limbs to be explicitly correlated in motor space, i.e., through intrinsic coordinates, we would have constrained the phases of the two hands/arms to be perfectly tuned.
This is not what happens in practice, as it is exemplified in the experiments by Heuer and colleagues \cite{Heueretal1998} and Byblow \textit{et al} \cite{Byblowetal2000}.
Since, in real-world gesture execution, perfect synchronisation seldom happens, a better result is obtained -- on average -- when we allow for the maximum flexibility in gesture execution.
Probability-based measures enforce such flexibility even more, because they do not constraint the distance metric to be tied to a specific time instant of the gesture execution. 

Finally, it is noteworthy that the use of Gaussian mixtures allows for determining what parts of the motion are more relevant, for those are characterised by lower amplitudes of the covariance.
As a consequence, the covariance matrix associated with each element in the model assures to give a proper weight to the sample itself. 
This allows to consider the differing importance of the various gesture's phases in the recognition phase, therefore taking into account motion dexterity and robustness. 

The combination of the two methods for modelling and recognition allows us to support the Bernstein's intuition about motion constraints \cite{Bernstein1967}, i.e., (i) variations in degrees of freedom affecting motion performance are constrained (in our case, by low covariance values in intra-arm correlation of inertial data); (ii) variation in degrees of freedom that \textit{do not affect} task performance can be unconstrained (again, by larger values in the covariance in intra-arm correlation); (iii) co-variations between gesture-relevant degrees of freedom not impacting on performance are permitted (by considering implicit correlations among the two hands/arms).

\section{Conclusions}
\label{sec:conclusions}

This paper discusses design choices involved in the detection and classification of bimanual gestures in unconstrained environments.
The assumption we make is the availability of inertial data originating from two distinct sensing points, conveniently located at the wrist, given the availability of such COTS sensors as smartwatches.
Our models are grounded with respect to Gaussian Mixture Modelling (GMM) and Gaussian Mixture Regression (GMR), which we use as basic modelling procedure.
Starting from these two positions, we compare different modelling (i.e., explicit and implicit correlations between the two hands/arms) and classification techniques (i.e., based on distance and probability-related considerations), which are inspired by literature about the representation of bimanual gestures in the brain. 
Our architecture allows for different combinations of modelling and classification techniques. 
Furthermore, it can be extended as a framework to support reproducible research.

Experiments show results related to $5$ generic bimanual activities, which have been selected on the basis of three main parameters: (not) constraining the two hands by a physical tool, (not) requiring a specific sequence of single-hand gestures, being recursive (or not). 
The best results are obtained when considering an implicit coordination among the two hands/arms (i.e., the two motions are modelled separately) and using a probability-based distance for classification (i.e., the specific timing characteristics of the trajectories are considered only to a limited extent). 
This seems to confirm a few insights from the literature related to motor control of bimanual gestures, and opens up a number of interesting research questions for the upcoming future.
\section*{References}

\bibliography{references}

\begin{thebibliography}{10}
\expandafter\ifx\csname url\endcsname\relax
  \def\url#1{\texttt{#1}}\fi
\expandafter\ifx\csname urlprefix\endcsname\relax\def\urlprefix{URL }\fi
\expandafter\ifx\csname href\endcsname\relax
  \def\href#1#2{#2} \def\path#1{#1}\fi

\bibitem{Kelso1984}
J.~Kelso, Phase transitions and critical behaviour in human bimanual
  coordination, American Journal of Physiology - Regulatory 15 (1984)
  R1000--R1004.

\bibitem{Hakenetal1985}
H.~Haken, J.~Kelso, H.~Bunz, A theoretical model of phase transitions in human
  hand movements, Biological Cybernetics 51 (1985) 347--356.

\bibitem{Wiles11}
J.~Wiles, A.~Leibing, N.~Guberman, J.~Reeve, R.~Allen, The meaning of ``ageing
  in place'' to older people, The gerontologist.

\bibitem{Mechsneretal2001}
F.~Mechsner, D.~Kerzel, G.~Knoblich, W.~Prinz, Perceptual basis of bimanual
  coordination, Nature 414 (2001) 69--72.

\bibitem{Sakuradaetal2015}
T.~Sakurada, K.~Ito, H.~Gomi, Bimanual motor coordination controlled by
  cooperative interactions in intrinsic and extrinsic coordinates, European
  Journal of Neuroscience 43 (2016) 120--130.

\bibitem{Heueretal1998}
H.~Heuer, W.~Spijkers, T.~Kleinserge, H.~van~der Loo, C.~Steglich, The time
  course fo cross-talk during the simultaneous specification of bimanual
  movement amplitudes, Experimental Brain Research 118 (1998) 381--392.

\bibitem{Byblowetal2000}
W.~Byblow, G.~Lewis, J.~Stinear, N.~Austin, M.~Lynch, The subdominant hand
  increases in the efficacy of voluntary alterations in bimanual coordination,
  Experimental Brain Research 131 (2000) 366--374.

\bibitem{SwinnenandWenderoth2004}
S.~Swinnen, N.~Wenderoth, Two hands, one brain: cognitive neuroscience of
  bimanual skill, Trends in Cognitive Science 8 (2004) 18--25.

\bibitem{Kodamaetal2015}
K.~Kodama, N.~Fukuyama, T.~Inamura, Differing dynamics of interpersonal and
  interpersonal coordination: two-finger and four-finger tapping experiments,
  PLOS one 10~(6).

\bibitem{BaltesandLindenberger1997}
P.~Baltes, U.~Lindenberger, Emergence of a powerful connection between sensory
  and cognitive functions across the adult life span: a new window to the study
  of cognitive ageing?, Psychological Ageing 12 (1997) 12--21.

\bibitem{SleimenMalkounetal2014}
R.~Sleimen-Malkoun, J.-J. Temprado, S.~L. Hong, Aging induced loss of
  complexity and dedifferentiation: consequences for coordination dynamics
  within and between brain, muscular and behavioural levels, Frontiers in
  Ageing Neuroscience 6 (2014) 1--17.

\bibitem{Rosenbaumetal2007}
D.~Rosenbaum, R.~Cohen, S.~Jax, R.~V. der Wel, D.~Weiss, The problem of serial
  order in behaviour: Lashley’s legacy, Human Movement Science 26 (2007)
  525--554.

\bibitem{Schacketal2014}
T.~Schack, K.~Essig, C.~Frank, D.~Koester, Mental representation and motor
  imagery training, Frontiers in Human Neuroscience 8 (2014) 1--10.

\bibitem{Schack2012}
T.~Schack, Measuring mental representations, in: G. Tenenbaum, R. Edlund, A.
  Kanata (Eds). Handbook of Measurement in Sport and Exercise Psychology, Human
  Kinetics, Urbana Champaign, IL, USA, 2012.

\bibitem{Bernstein1967}
N.~Bernstein, The coordination and regulation of movements, Pergamon Press,
  Oxford, 1967.

\bibitem{Bruno12}
B.~Bruno, F.~Mastrogiovanni, A.~Sgorbissa, T.~Vernazza, R.~Zaccaria, Human
  motion modeling and recognition: a computational approach, in: Proceedings of
  the 8th IEEE International Conference on Automation Science and Engineering
  (CASE 2012), Seoul, Korea, 2012.

\bibitem{Lara13}
O.~Lara, M.~Labrador, A survey on human activity recognition using wearable
  sensors, IEEE Communications Surveys and Tutorials 15~(3) (2013) 1192--1209.

\bibitem{Lester05}
J.~Lester, T.~Choudhury, N.~Kern, G.~Borriello, B.~Hannaford, A hybrid
  discriminative/generative approach for modeling human activities, in:
  Proceedings of the 19th International Joint Conference on Artificial
  Intelligence (IJCAI 2005), Edinburgh, UK, 2005.

\bibitem{Mathie04}
M.~Mathie, B.~Celler, N.~Lovell, A.~Coster, Classification of basic daily
  movements using a triaxial accelerometer, Medical \& Biological Engineering
  \& Computing 42 (2004) 679--687.

\bibitem{Dietrich14}
M.~Dietrich, K.~von Laerhoven, Recall your actions! using wearable activity
  recognition to augment the human mind, in: Proceedings of the 2014 ACM
  International Joint Conference on Pervasive and Ubiquitous Computing (UbiComp
  2014), Seattle, Washington, USA, 2014.

\bibitem{Bulling14}
A.~Bulling, U.~Blanke, B.~Schiele, A tutorial on human activity recognition
  using body-worn inertial sensors, ACM Computing Surveys 46~(3) (2014)
  33:1--33:33.

\bibitem{Krassnig10}
G.~Krassnig, D.~Tantinger, C.~Hofmann, T.~Wittenberg, M.~Struck, User-friendly
  system for recognition of activities with an accelerometer, in: Proceedings
  of the 2010 International Conference on Pervasive Computing Technologies for
  Healthcare (PervasiveHealth 2010), Munich, Germany, 2010.

\bibitem{Mashita12}
T.~Mashita, K.~Shimatani, M.~Iwata, H.~Miyamoto, D.~Komaki, T.~Hara,
  K.~Kiyokawa, H.~Takemura, S.~Nishio, Human activity recognition for a content
  search system considering situations of smartphone users, in: Proceedings of
  the IEEE Workshop on Virtual Reality Short Papers and Posters (VRW 2012),
  Orange County, California, USA, 2012.

\bibitem{Lee11}
M.~Lee, A.~Khan, T.~Kim, A single tri-axial accelerometer-based real-time
  personal life log system capable of human activity recognition and exercise
  information generation, Personal and Ubiquitous Computing 15~(8) (2011)
  887--898.

\bibitem{Karantonis06}
D.~Karantonis, M.~Narayanan, M.~Mathie, N.~Lovell, B.~Celler, Implementation of
  a real-time human movement classifier using a triaxial accelerometer for
  ambulatory monitoring, IEEE Transactions on Information Technology in
  Biomedicine 10~(1) (2006) 156--167.

\bibitem{Chul09}
H.~Chul, P.~Kiheon, Estimating accelerated body attitude using an inertial
  sensor, in: Proceedings of the 2009 International Joint Conference ICROS-SICE
  (ICROS-SICE 2009), Fukuoka, Japan, 2009.

\bibitem{Bonnet07}
S.~Bonnet, R.~Heliot, A magnetometer-based approach for studying human
  movements, IEEE Transactions on Biomedical Engineering 54~(7) (2007)
  1353--1355.

\bibitem{Sharma08}
A.~Sharma, A.~Purwar, Y.~Lee, Y.~Lee, W.~Chung, Frequency based classification
  of activities using accelerometer data, in: Proceedings of the 2008
  International Conference on Multisensor Fusion and Integration for
  Intelligent Systems (MFI 2008), Seoul, Korea, 2008.

\bibitem{Cho08}
Y.~Cho, Y.~Nam, Y.~Choi, W.~Cho, Smartbuckle: Human activity recognition using
  a 3-axis accelerometer and a wearable camera, in: Proceedings of the 2nd
  International Workshop on Systems and Networking Support for Healthcare and
  Assisted Living Environments (HealthNet 2008), Breckenridge, Colorado, USA,
  2008.

\bibitem{Lee02}
S.~Lee, K.~Mase, Activity and location recognition using wearable sensors, IEEE
  Pervasive Computing 1~(3) (2002) 24--32.

\bibitem{Minnen05}
D.~Minnen, T.~Starner, J.~Ward, P.~Lukowicz, G.~Troster, Recognizing and
  discovering human actions from on-body sensor data, in: Proceedings of the
  2005 {IEEE} International Conference on Multimedia and Expo (ICME 2005),
  Amsterdam, The Netherlands, 2005.

\bibitem{OlguinOlguin06}
D.~Olguin, A.~Pentland, Human activity recognition: accuracy across common
  locations for wearable sensors, in: Proceedings of the 2006 {IEEE}
  International Symposium on Wearable Computers (ISWC 2006), Montreaux,
  Switzerland, 2006.

\bibitem{Choudhury08}
T.~Choudhury, S.~Consolvo, B.~Harrison, J.~Hightower, A.~LaMarca, L.~LeGrand,
  A.~Rahimi, A.~Rea, G.~Borriello, B.~Hemingway, P.~Klasnja, K.~Koscher,
  J.~Landay, J.~Lester, D.~Wyatt, D.~Haehnel, The mobile sensing platform: an
  embedded activity recognition system, IEEE Pervasive Computing Magazine 7~(2)
  (2008) 32--41.

\bibitem{Bruno13}
B.~Bruno, F.~Mastrogiovanni, A.~Sgorbissa, T.~Vernazza, R.~Zaccaria, Analysis
  of human behaviour recognition algorithms based on acceleration data, in:
  Proceedings of the 2013 IEEE International Conference on Robotics and
  Automation (ICRA 2013), Karlsruhe, Germany, 2013.

\bibitem{Calinon10}
S.~Calinon, F.~D'Halluin, E.~Sauser, D.~Caldwell, A.~Billard, Learning and
  reproduction of gestures by imitation, IEEE Robotics \& Automation Magazine 6
  (2010) 44--54.

\bibitem{Rousseeuw87}
P.~Rousseeuw, Silhouettes: a graphical aid to the interpretation and validation
  of cluster analysis, Journal of Computational and Applied Mathematics 20
  (1987) 53--65.

\bibitem{Mahalanobis36}
P.~Mahalanobis, On the generalized distance in statistics, Proceedings of the
  National Institute of Sciences of India 2~(1) (1936) 49--55.

\bibitem{Bruno14b}
B.~Bruno, F.~Mastrogiovanni, A.~Saffiotti, A.~Sgorbissa, Using fuzzy logic to
  enhance classification of human motion primitives, in: Proceedings of the
  15th International Conference on Information Processing and Management of
  Uncertainty in Knowledge-Based Systems (IPMU 2014), Montpellier, France,
  2014.

\end{thebibliography}

\end{document}